\documentclass{article}

\usepackage{microtype}
\usepackage{graphicx}
\usepackage{booktabs} 

\usepackage{hyperref}

\usepackage[accepted]{icml2024}

\usepackage{amsmath}
\usepackage{amssymb}
\usepackage{mathtools}
\usepackage{amsthm}

\usepackage[capitalize,noabbrev]{cleveref}

\theoremstyle{plain}
\newtheorem{theorem}{Theorem}[section]
\newtheorem{proposition}[theorem]{Proposition}
\newtheorem{lemma}[theorem]{Lemma}
\newtheorem{corollary}[theorem]{Corollary}
\theoremstyle{definition}

\theoremstyle{remark}

\usepackage[textsize=tiny]{todonotes}

\usepackage{amsmath,amsfonts,bm}

\def\eqref#1{equation~\ref{#1}}

\def\1{\bm{1}}

\def\eps{{\epsilon}}

\def\vh{{\bm{h}}}

\def\vx{{\bm{x}}}
\def\vy{{\bm{y}}}

\def\mA{{\bm{A}}}

\def\mD{{\bm{D}}}

\def\mH{{\bm{H}}}
\def\mI{{\bm{I}}}

\def\mM{{\bm{M}}}

\def\mR{{\bm{R}}}

\def\mW{{\bm{W}}}
\def\mX{{\bm{X}}}
\def\mY{{\bm{Y}}}

\DeclareMathAlphabet{\mathsfit}{\encodingdefault}{\sfdefault}{m}{sl}
\SetMathAlphabet{\mathsfit}{bold}{\encodingdefault}{\sfdefault}{bx}{n}

\def\gG{{\mathcal{G}}}

\def\gS{{\mathcal{S}}}

\def\sA{{\mathbb{A}}}

\def\sL{{\mathbb{L}}}

\def\sR{{\mathbb{R}}}
\def\sS{{\mathbb{S}}}

\def\sV{{\mathbb{V}}}

\newcommand{\E}{\mathbb{E}}

\usepackage{url}

\usepackage{subcaption}

\newcommand*{\scale}[2][4]{\scalebox{#1}{$#2$}}%

\newcommand{\ttest}[0]{\textit{t}-test}

\newcommand{\MLP}[0]{\mathrm{MLP}}

\newcommand{\clamp}[0]{\mathrm{clamp}}

\newcommand{\where}[0]{\text{where}}

\newcommand{\mean}[0]{\text{mean}}
\newcommand{\std}[0]{\text{std}}
\newcommand{\diag}[0]{\text{diag}}
\newcommand{\node}[0]{\text{node}}

\newcommand{\readout}[0]{\texttt{READOUT}}
\newcommand{\gnn}[0]{\texttt{GNN}}
\newcommand{\enc}[0]{\texttt{ENC}}

\newcommand{\txtrain}{\text{train}}

\newcommand{\txeval}{\text{eval}}

\newcommand{\hE}[0]{\hat{E}}
\newcommand{\hv}[0]{\hat{v}}
\newcommand{\he}[0]{\hat{e}}
\newcommand{\hd}[0]{\hat{d}}
\newcommand{\hgG}[0]{\hat{\gG}}
\newcommand{\hsV}[0]{\hat{\sV}}
\newcommand{\hmX}[0]{\hat{\mX}}
\newcommand{\hmA}[0]{\hat{\mA}}

\newcommand{\hmD}[0]{\hat{\mD}}

\newcommand{\hvh}[0]{\hat{\vh}}
\newcommand{\hvx}[0]{\hat{\vx}}
\newcommand{\hvy}[0]{\hat{\vy}}
\newcommand{\hmH}[0]{\hat{\mH}}
\newcommand{\hmY}[0]{\hat{\mY}}

\definecolor{weakgray}{HTML}{e7e7e7}
\definecolor{weakblue}{HTML}{E3F2FD}
\definecolor{weakred}{HTML}{FFEBEE}
\definecolor{midblue}{HTML}{bbdefb}
\definecolor{midred}{HTML}{ffcdd2}

\definecolor{darkblue}{HTML}{1565c0}
\definecolor{darkred}{HTML}{c62828}

\newcommand{\STON}[0]{\text{S2N}}

\newcommand{\sub}[0]{\text{sub}}

\newcommand{\co}[0]{\text{co}}

\newcommand{\PPIBP}[0]{\textsf{\fontsize{8.5pt}{8.5pt}\selectfont PPI-BP}}
\newcommand{\HPOMetab}[0]{\textsf{\fontsize{8.5pt}{8.5pt}\selectfont HPO-Metab}}
\newcommand{\HPONeuro}[0]{\textsf{\fontsize{8.5pt}{8.5pt}\selectfont HPO-Neuro}}
\newcommand{\EMUser}[0]{\textsf{\fontsize{8.5pt}{8.5pt}\selectfont EM-User}}

\newcommand{\PPIBPb}[0]{\textsf{\fontsize{8.5pt}{8.5pt}\selectfont PPI-BP} }

\newcommand{\HPONeurob}[0]{\textsf{\fontsize{8.5pt}{8.5pt}\selectfont HPO-Neuro} }
\newcommand{\EMUserb}[0]{\textsf{\fontsize{8.5pt}{8.5pt}\selectfont EM-User} }

\newcommand{\sHPONeuro}[0]{\textsf{\fontsize{7.25pt}{7.25pt}\selectfont HPO-Neuro}}

\newcommand{\Density}[0]{\textsf{\fontsize{8.5pt}{8.5pt}\selectfont Density}}
\newcommand{\CutRatio}[0]{\textsf{\fontsize{8.5pt}{8.5pt}\selectfont Cut-Ratio}}
\newcommand{\Coreness}[0]{\textsf{\fontsize{8.5pt}{8.5pt}\selectfont Coreness}}
\newcommand{\Component}[0]{\textsf{\fontsize{8.5pt}{8.5pt}\selectfont Component}}

\newcommand{\Densityb}[0]{\textsf{\fontsize{8.5pt}{8.5pt}\selectfont Density} }
\newcommand{\CutRatiob}[0]{\textsf{\fontsize{8.5pt}{8.5pt}\selectfont Cut-Ratio} }

\newcommand{\propositionSingleGCNApproxContent}[0]{
Using the single-layer GCN parametrized by $\mW$, subgraph representations $\mR^{\top}\mH$ of the global graph $\gG$ can be approximated by node representations $\hmH$ of the S2N graph $\hgG$, that is, $\hmH \approx \mR^{\top}\mH$. The error between two representations is bounded by:
\vspace{-0.13cm}
\begin{align}\label{eq:single_gcn_approx}
\lVert \mR^{\top}\mH - \hmH \rVert
    \leq
    M^{\frac{1}{2}}
    \lVert
        \mX - \mR \hmX
    \rVert
        \cdot
    \lVert \mW \rVert.
\end{align}
}

\newcommand{\propositionTimeComplexityContent}[0]{
The time complexity of the 1-layer GLASS, Connected form, S2N+0, and S2N+A is
\vspace{-0.2cm}
{\small
\begin{align}
\text{GLASS \& Connected: }& O ( EF + M \overline{N^{\sub}} F + NF^2 ), \\
\text{S2N+0: }& O ( \hE F +  M \overline{N^{\sub}} F + MF^2  ), \\
\text{S2N+A: }& O ( \hE F +  M \overline{E^{\sub}} F + M \overline{N^{\sub}} F^2  ).
\end{align}}
}

\newcommand{\propositionEdgeCMContent}[0]{
For Configuration Model as $\gG$ and i.i.d. sampled subgraphs where the average size is $\overline{N^\sub}$, the probability that the weight $\hmA_{[i, j]}$ of an edge $(i, j)$ in $\hgG$ is bigger than $c > 0$ is $P(\hmA_{[i, j]} \geq c) \leq \frac{ (\overline{N^\sub} )^2 \E[d]  }{cN}$ where $\E[d]$ is an average degree of $\gG$.
}

\newcommand{\propositionSingleGCNApprox}[0]{
\begin{proposition}\label{proposition:single_gcn_approx}
\propositionSingleGCNApproxContent
\end{proposition}
}

\newcommand{\propositionTimeComplexity}[0]{
\begin{proposition}\label{proposition:time_complexity}
\propositionTimeComplexityContent
\end{proposition}
}

\newcommand{\propositionEdgeCM}[0]{
\begin{proposition}\label{proposition:edge_cm}
\propositionEdgeCMContent
\end{proposition}
}

\usepackage{lipsum}
\usepackage{CJKutf8}
\usepackage{multirow}
\usepackage{colortbl}
\usepackage{soul}
\usepackage{tabularx}
\usepackage{wrapfig}

\icmltitlerunning{Translating Subgraphs to Nodes Makes Simple GNNs Strong and Efficient for Subgraph Representation Learning}

\begin{document}

\twocolumn[
\icmltitle{Translating Subgraphs to Nodes Makes Simple GNNs \\ Strong and Efficient for Subgraph Representation Learning}



\icmlsetsymbol{equal}{*}

\begin{icmlauthorlist}
\icmlauthor{Dongkwan Kim}{kaist}
\icmlauthor{Alice Oh}{kaist}
\end{icmlauthorlist}

\icmlaffiliation{kaist}{School of Computing, KAIST, South Korea}

\icmlcorrespondingauthor{Dongkwan Kim}{dongkwan.kim@kaist.ac.kr}
\icmlcorrespondingauthor{Alice Oh}{alice.oh@kaist.edu}

\icmlkeywords{Machine Learning, ICML}

\vskip 0.3in
]



\printAffiliationsAndNotice{}  

\begin{abstract}
Subgraph representation learning has emerged as an important problem, but it is by default approached with specialized graph neural networks on a large global graph. These models demand extensive memory and computational resources but challenge modeling hierarchical structures of subgraphs. In this paper, we propose Subgraph-To-Node (S2N) translation, a novel formulation for learning representations of subgraphs. Specifically, given a set of subgraphs in the global graph, we construct a new graph by coarsely transforming subgraphs into nodes. Demonstrating both theoretical and empirical evidence, S2N not only significantly reduces memory and computational costs compared to state-of-the-art models but also outperforms them by capturing both local and global structures of the subgraph. By leveraging graph coarsening methods, our method outperforms baselines even in a data-scarce setting with insufficient subgraphs. Our experiments on eight benchmarks demonstrate that fined-tuned models with S2N translation can process 183 -- 711 times more subgraph samples than state-of-the-art models at a better or similar performance level.
\end{abstract}

\section{Introduction}\label{sec:introduction}
Subgraph representation learning has been shown to be useful for various real-world problems~\citep{bordes2014question,luo2022shine,hamidi2022subgraph,maheshwari2024timegraphs}. Current research uses the default data structures for graph-level tasks, treating the subgraph as just a subset of the global graph. Existing studies on subgraph representation learning focus on developing graph neural networks specialized for subgraphs~\citep{alsentzer2020subgraph, wang2022glass}. However, specialized models suffer from large memory and computational requirements from a large global graph. Their complex operations also lead to limitations in learning both local and global interactions of subgraphs. Here, we address a fundamental yet underexplored question in subgraph representation learning before model design: \textbf{How can we efficiently and effectively process subgraphs as data for representation learning?}

In this paper, we propose `Subgraph-To-Node (S2N)', a novel data structure to solve subgraph-level prediction tasks efficiently. This data structure is a new graph translated from the original global graph and subgraphs, where its nodes are the original subgraphs, and its edges are the relations among the original subgraphs. Then, we can get the results of the subgraph-level tasks by performing node-level tasks from these node representations.

For example, \citet{alsentzer2020subgraph} introduces a fitness social network where subgraphs are users, nodes are workouts, and edges indicate if multiple users complete workouts. By using S2N translation, a new graph is created; users become nodes, and edges express relations between them. This graph directly shows the relationships between users, following the conventional approach of describing social networks where nodes are users. As seen in this example, S2N can provide a more precise description of real-world problems than a form of subgraphs.

The S2N translation enables efficient subgraph representation learning. The number of nodes in the S2N graph is decreased to the number of original subgraphs. The edges of the S2N graph are also significantly reduced, which we theoretically prove and empirically confirm in real-world datasets. We can load large batches of subgraphs on the GPU memory and parallelize the training and inference. Since S2N translation does not interfere with model selection, even simple GNNs without complex operations can encode node representations in the S2N graph.

There can be various implementations of S2N translation, and here, we create new edges as the number of shared edges across a pair of subgraphs. Then, we normalize and sparsify edges based on weights to approximate the structure of the global graph. This process makes a coarse S2N graph carry sufficient information across subgraphs for the task. We can additionally preserve structural information during S2N translation using two strategies. First, we propose S2N that retains internal structures in subgraphs. Second, we incorporate structural encoding into input node features~\citep{dwivedi2022graph}. These enhancements require negligible resources, as they are performed only once before training and are efficiently performed with low complexity of computation and space.

Furthermore, we address S2N's challenge when there are not sufficient samples available, specifically, representing parts of the global graph not covered by existing subgraphs. We introduce Coarsened S2N (CoS2N), which uses graph coarsening to create `virtual' subgraphs that summarize the global structure. The CoS2N allows message-passing between distant subgraphs with labels without compromising efficiency. We also theoretically show that CoS2N can reduce the approximation error in S2N's representations.

We conduct experiments with four real-world and four synthetic datasets to evaluate the classification performance as well as efficiency, including throughput, latency, parameters, and memory usage. We demonstrate that models using S2N transformation are more efficient than existing approaches, with similar or even better performance. Specifically, while best-tuned models with S2N can process $183$ -- $711$ times as many samples, their classification performance shows $99.9$ -- $102.9$\% of the state-of-the-art model.

The rest of the paper is organized as follows. First, we present a Subgraph-To-Node (S2N) translation, a novel way to generate an efficient data structure for subgraph representation learning (\S\ref{sec:method}). This section includes Coarsened S2N (CoS2N), the combination with graph coarsening to tackle a data-scarce setting. Second, we theoretically show that S2N reduces the computational complexity and approximates subgraph representations from the original global graph (\S\ref{sec:theoretical_analysis}). Third, we demonstrate the efficiency of S2N compared to the state-of-the-art approaches, specifically enabling up to 711 times the throughput while maintaining the performance of at least 99.9\% (\S\ref{sec:experiments}, \S\ref{sec:results}).

\section{Related Work}\label{sec:related_work}
Our S2N translation tackles representation learning of subgraphs, and this is closely linked to graph coarsening. We introduce these fields and their connection with our study. We discuss more related work in Appendix~\ref{appendix:related_work}.

\vspace{-0.43cm}
\paragraph{Subgraph Representation Learning}

Learning subgraph representations has been beneficial across various real-world problems. Researchers model higher-order interactions by subgraphs that nodes, edges, and graphs cannot. For example, diseases and patients in gene networks~\citep{luo2022shine}, teams in collaboration networks~\citep{hamidi2022subgraph}, and communities in mobile game user networks~\citep{zhang2023constrained} are represented by subgraphs. However, they are specialized for each domain~\citep{zhang2023constrained,li2023self,trumper2023performance,ouyang2024bitcoin,maheshwari2024timegraphs} or have strong assumptions about the subgraph~\citep{meng2018subgraph,hamidi2022subgraph,kim2022models,luo2022shine}, making them difficult to generalize. The Subgraph Neural Network (SubGNN)~\citep{alsentzer2020subgraph} is the first general approach to subgraph representation learning using topology, positions, and connectivity. The GNN with LAbeling trickS for Subgraph (GLASS)~\citep{wang2022glass} uses a labeling trick to distinguish nodes inside and outside the subgraph and enhance the expressive power of representations. However, both SubGNN and GLASS perform complex operations on a large global graph, demanding high memory and computation but not reflecting the layered structures of subgraphs. Our method allows efficient learning of subgraph representations without a complex model design.

\vspace{-0.22cm}
\paragraph{Graph Coarsening}

Our S2N translation summarizes subgraphs into nodes, and in that sense, it is related to graph coarsening (or summarization) methods~\citep{loukas2018spectrally, loukas2019graph, Deng2020GraphZoom:, cai2021graph, huang2021scaling, zhou2021dpgs, jin2022graph}. These methods are similar to our work that aims to handle large-scale graphs efficiently, but they have not been applied to subgraph-level tasks. Moreover, the super-nodes in coarse graphs are not given to existing graph coarsening methods; thus, algorithms to decide on super-nodes are required. In S2N translation, we treat subgraphs as super-nodes and can create coarse graphs with nominal costs.

\section{Data Structures for Subgraph Representation Learning}\label{sec:method}

\begin{figure*}[t]
  \centering
  \begin{subfigure}[b]{0.42\textwidth}
    \centering
    \includegraphics[width=0.96\textwidth]{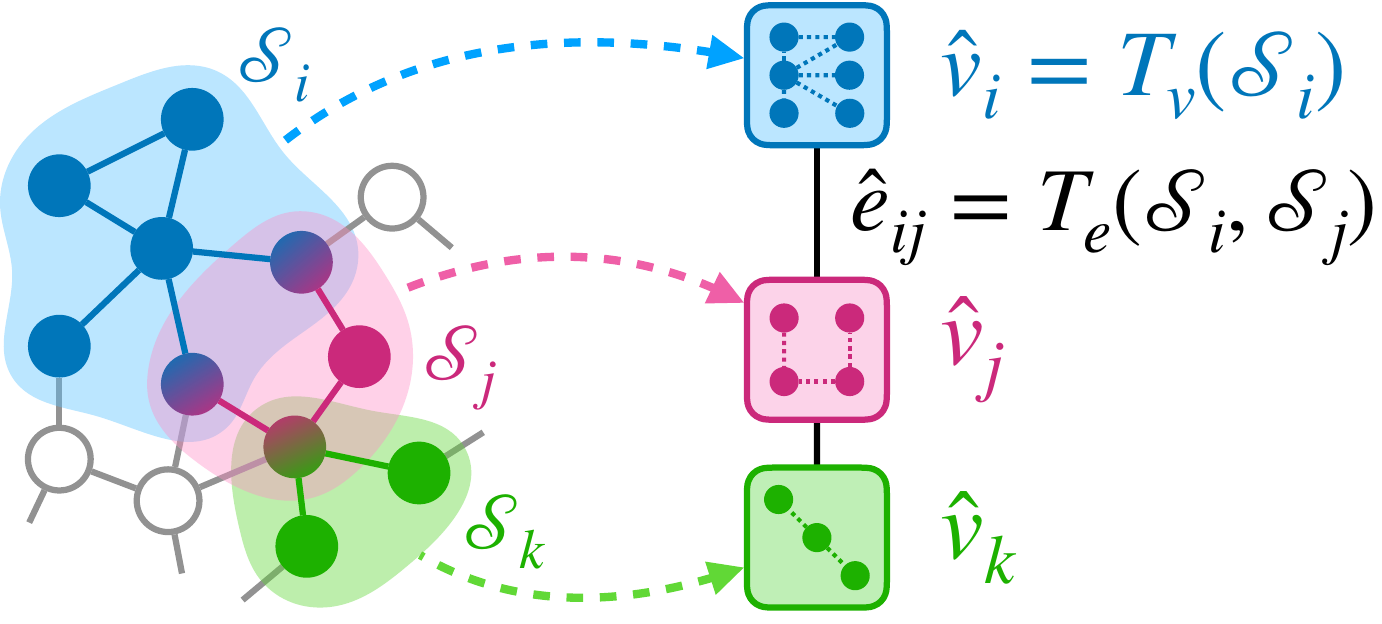}
    \vspace{0.2cm}
    \caption{The S2N translation. Subgraphs $\gS_i$ and $\gS_j$ are transformed into nodes $\hv_i$ and $\hv_j$ by $T_v$, and an edge $\he_{ij}$ between them is formed by $T_e$.}
    \label{fig:model_s2n}
  \end{subfigure}
     \hfill
  \begin{subfigure}[b]{0.515\textwidth}
    \centering
    \includegraphics[width=0.98\textwidth]{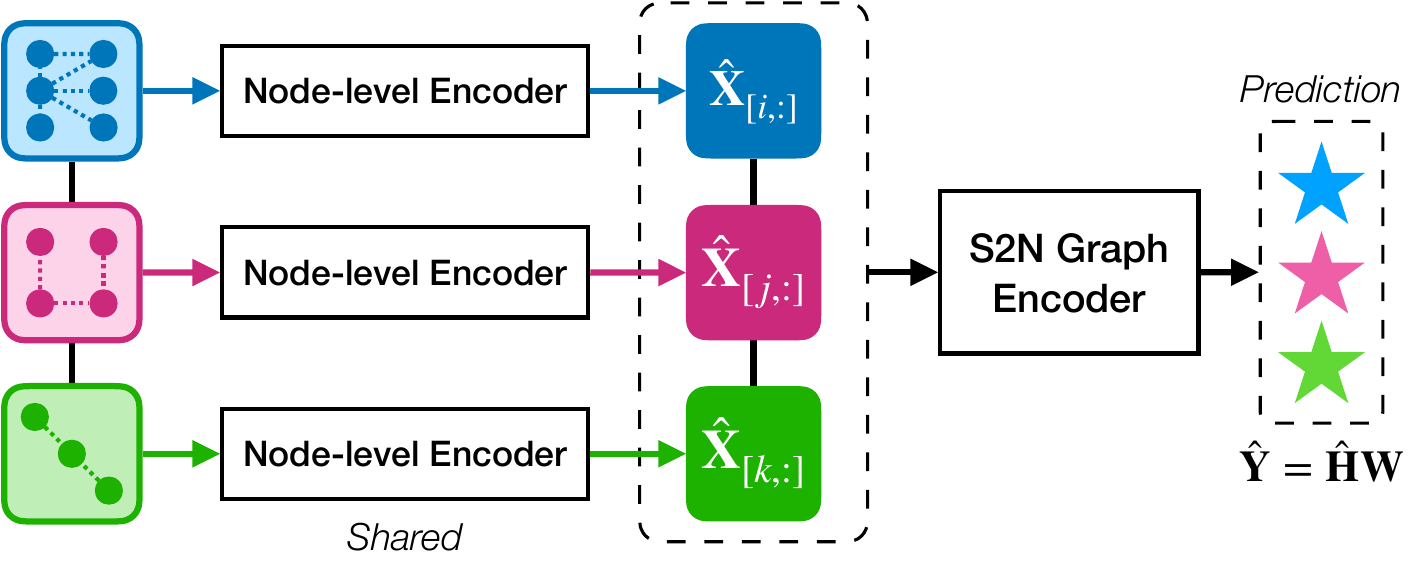}
    \vspace{0.1cm}
    \caption{Models for graphs translated by S2N. We apply a node-level encoder first (weighted sum for S2N+0 and GNN plus readout for S2N+A), then an S2N graph encoder (GNN) to their outputs for the prediction.}
    \label{fig:model_pipeline}
  \end{subfigure}
  \caption{Overview of the Subgraph-To-Node (S2N) translation and models for translated graphs.}
  \label{fig:model}
\end{figure*}

We introduce three data structures for subgraph representation learning including our proposed Subgraph-To-Node (S2N) translation.

\vspace{-0.2cm}
\paragraph{Notations}

We first summarize the notations in the subgraph representation learning for classification. Let $\gG = (\sV, \mA, \mX)$ be a global graph where $\sV$ is a set of nodes ($|\sV| = N$), $\mA \in \{ 0, 1 \}^{N \times N}$ is an adjacency matrix, and $\mX \in \sR^{N \times F_0}$ is a node feature matrix. A subgraph $\gS = (\sV^{\sub}, \mA^{\sub})$ is a graph formed by subsets of nodes and edges in the global graph $\gG$. For the subgraph classification task, there is a set of $M (< N)$ subgraphs $\sS = \{ \gS_1, \gS_2, ..., \gS_M \}$, and for $\gS_i = (\sV_i^{\sub}, \mA_i^{\sub})$, the goal is to learn its representation $\vh_i \in \sR^{F}$ and the logit vector $\vy_i \in \sR^{C}$ where $F$ and $C$ are the numbers of hidden features and classes, respectively.

\vspace{-0.1cm}
\subsection{Conventional Data Structures}
\vspace{-0.1cm}

The existing GNN-based approach employs two types of data structures when solving subgraph-level tasks. This paper refers to these two as Separated and Connected forms. The \textit{Separated} form treats each subgraph as a separate graph, applying the GNN instance-wise for each graph. 
Existing studies express these separated graphs as \textit{standalone} or \textit{segregated} graphs and use this separated form as the main baseline. The \textit{Connected} form represents subgraphs by applying the GNN on the whole global graph and pooling node representations. The separated form preserves only internal structures, and the connected form retains all information in the global graph. For this reason, using the connected form requires more memory and computational resources. Since incorporating the structures in the global graph is essential in learning subgraphs, we design a new data structure that can approximate the global graph without significant costs.

\subsection{Subgraph-To-Node (S2N) Translation}

The S2N translation reduces memory and computational costs by constructing a new coarse graph that summarizes the original subgraph into a node. As in Figure~\ref{fig:model_s2n}, for each subgraph $\gS_i \in \sS$ in the global graph $\gG$, we create a node $\hv_i = T_v(\gS_i)$ in the translated graph $\hgG$; for all pairs $(\gS_i, \gS_j)$ of subgraphs in $\gG$, we assign an edge $\he_{ij} = T_e (\gS_i, \gS_j)$ between corresponding nodes in $\hgG$. Here, $T_v$ and $T_e$ are translation functions for nodes and edges in $\hgG$, respectively. Formally, the S2N translated graph $\hgG = (\hsV, \hmA)$ where $|\hsV| = M$ and $\hmA \in \sR^{M \times M}$, is defined by
\vspace{-0.1cm}
\begin{equation}
\hsV = \{ T_{v} (\gS_{i})\ |\  \gS_{i} \in \sS \},\ 
\hmA_{[i,j]} = T_{e} (\gS_{i}, \gS_{j}).
\end{equation}
We can choose any function for $T_v$ and $T_e$. For example, $T_e$ can be simple heuristics (e.g., the distance between subgraphs) or modeled with neural networks to learn the graph structure~\citep{franceschi2019learning, kim2021how, fatemi2021slaps}.

In this paper, we choose two versions of S2N functions with negligible costs: \textbf{S2N+0} and \textbf{S2N+A}. For both versions, we use the same $T_e$ to make an edge and its weight as the number of edges between two subgraphs $\gS_i$ and $\gS_j$, which is defined as follows:
\vspace{-0.1cm}
\begin{equation}\label{eq:s2n_implementation}
T_{e} (\gS_{i}, \gS_{j})
= \textstyle \sum_{v_i \in \sV_i^{\sub}} \sum_{v_j \in \sV_j^{\sub}} \mA_{[v_i, v_j]}.
\end{equation}
When using edge weights as input, if the range of the values is too wide, learning may be unstable. So, we normalize and clamp the edge weights to between 0 to 1 by selecting edges in a specific range of standard scores ($a$ -- $b$ where $a, b$ are hyperparameters).
\begin{equation}\begin{aligned}\label{eq:s2n_edge_normalization}
\scale[0.94]{
\mathrm{normalize}(\hmA) = \clamp \textstyle \left(
\frac{  (\hmA - \mean(\hmA))  / \std(\hmA) - a}{b - a}
\right)
} \\
\scale[0.94]{
\ \where \ \ 
\clamp (x) = \max \left(0, \min \left(1,
x
\right) \right).
}
\end{aligned}\end{equation}
For $T_v$, we use different functions for S2N+0 and S2N+A. The difference between the two is whether it maintains the internal structures $\mA^{\sub}_i$ of the subgraph $\gS_i = (\sV^{\sub}_i, \mA^{\sub}_i)$. S2N+0 uses $T_v$ that ignores $\mA^{\sub}_i$ and treats the node as a set (i.e., $\sV^{\sub}_i$). In contrast, S2N+A's $T_v$ retains all information of nodes and edges in the subgraph:
\begin{equation}\begin{aligned}
\scale[0.92]{
\textbf{S2N+0:  }
    T_{v} (\gS_{i}) = \sV_i^{\sub},
    \ 
\textbf{S2N+A:  }
    T_{v} (\gS_{i}) = (\sV_{i}^{\sub}, \mA_{i}^{\sub}).
}
\end{aligned}\end{equation}
Note that their names originated from whether the adjacency matrix is a zero matrix ($0$) or not ($A$).

We can enhance the input features $\mX$ by incorporating structural encoding, thereby preserving more information when S2N summarizes global structures. In this paper, we adopt Random Walk Positional Encoding (RWPE)~\citep{dwivedi2022graph} that encodes the $k$-hop topology of the global graph for each node. The efficiency of S2N is maintained since RWPE is computed once before training and only requires the space complexity of $O(N)$. 

\subsection{Models for S2N Translated Graphs}

We propose simple but strong models for S2N (Figure~\ref{fig:model_pipeline}): node-level encoder $\enc_{\node}$ and S2N graph encoder $\enc_{\STON}$. First, $\enc_{\node}$ takes $T_v(\gS)$ as an input and produces $\hvx_i \in \sR^{F}$, input vector for the node in the S2N graph. Then, $\enc_{\STON}$ takes $\hmX = [ \hvx_1, ..., \hvx_M ]^{\top} \in \sR^{M \times F}$ and $\hmA$ as inputs, and produces representations $\hmH = [ \hvh_1, ..., \hvh_M ]^{\top} \in \sR^{M \times F}$ and logits $\hmY = [ \hvy_1, ..., \hvy_M ]^{\top} \in \sR^{M \times C}$.

For $\enc_{\node}$, we use different models for S2N+0 and S2N+A. Since the node in S2N+0 is a set of original nodes in $\gS_i$, we take a set of node features in $\sV_i$ as an input and generate a weighted sum of them. For S2N+A, we apply a GNN model to each subgraph as an individual graph, then apply a weighted sum for readout. Formally,
\begin{equation}\begin{aligned} 
&\scale[0.92]{
    \textbf{S2N+0:  }
    \hvx_i \textstyle
        = \sum_{v \in \sV_i^{\sub}} \omega_{vi} \cdot \mX_{[v, :]},
}
    \\
&\scale[0.9]{
\textbf{S2N+A:  }
    \hvx_i \textstyle
        = \sum_{v \in \sV_i^{\sub}} \omega_{vi} \cdot \gnn_{\node} ( \mX_{[\sV_i^{\sub}, :]}, \mA_{i}^{\sub})_{[v, :]},
}
\end{aligned}\end{equation}
where $\omega_{vi}$ is a weight corresponding to the node $v$ and the subgraph $\gS_i$. These weights can be either learnable or constants (e.g., $\omega = 1$ means that $\hvx$ is the sum of features). 

Given $\hmA$ and $\hmX$ of S2N+0 and S2N+A, we apply the S2N graph encoder $\enc_{\STON}$ which is another $\gnn_{\STON}$ to generate the final node representations $\hmH$ and logits $\hmY$ for prediction, That is, $\hmH = \gnn_{\STON} (\hmX, \hmA)$ and $\hmY = \hmH \mW$ where $\mW \in \sR^{F \times C}$ is a matrix of parameters. We can take any GNNs that perform message-passing between nodes. This node-level message-passing on translated graphs is analogous to message-passing at the subgraph level in SubGNN~\citep{alsentzer2020subgraph}.

\subsection{Coarsened S2N for a Data-Scarce Setting}\label{sec:coarsening}

\begin{figure}[t]
    \begin{center}
        \includegraphics[width=0.45\textwidth]{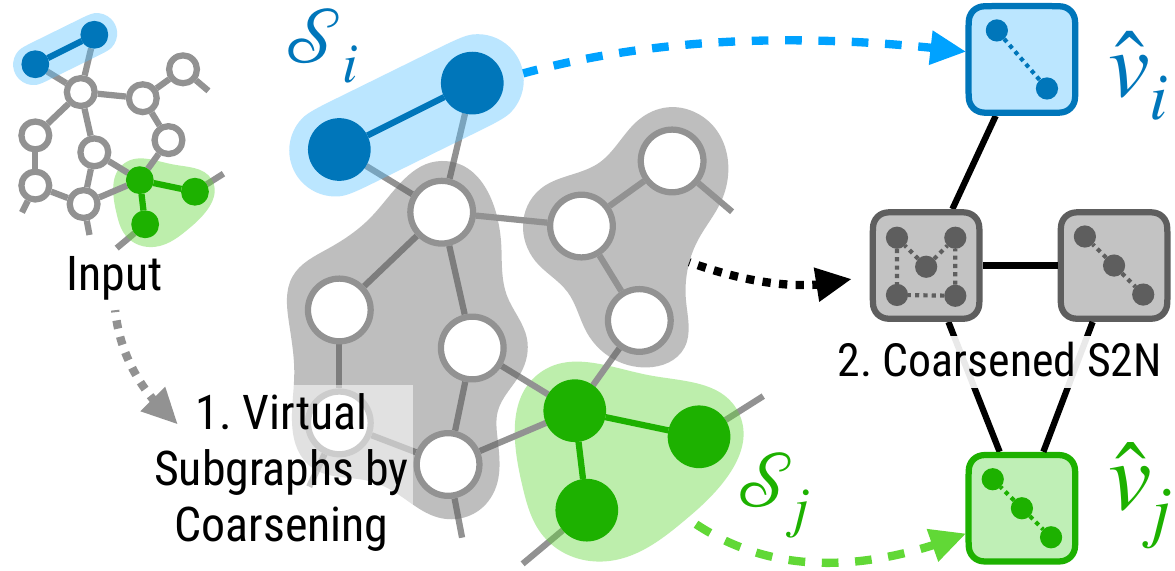}
    \end{center}
    \vspace{-0.1cm}
    \caption{Overview of Coarsened Subgraph-To-Node (CoS2N) Translation with virtual subgraphs generated by graph coarsening.}
    \label{fig:coarsening}
    \vspace{-0.1cm}
\end{figure}


By design, the S2N graph $\hgG$ can approximate the global graph $\gG$ covered by subgraphs, but cannot reflect parts of $\gG$ where subgraphs do not exist. When a pair of subgraphs is distant on the global graph, they exist as unconnected nodes in S2N graphs as illustrated in Figure~\ref{fig:coarsening}. These isolated subgraphs are likely to occur when the subgraph samples are scarce. In this case, GNNs cannot exchange supervised signals between subgraphs.

To solve this problem, we apply graph coarsening methods to the global graph $\gG$ to generate a partition of nodes in $\gG$. That is, graph coarsening summarizes a graph $\gG$ by grouping its nodes into super-nodes. Each node in $\gG$ corresponds to one super-node in the new graph $\hgG$. We construct induced subgraphs $\sS^{\co} = \{ \gS_1^{\co}, \gS_2^{\co}, ..., \gS_{M^\co}^{\co} \}$ for each super-node in the global graph.
Here, we call them `virtual subgraphs.' Using the original (labeled) subgraphs $\sS$ as is, the virtual subgraphs are merged with $\sS$ to form the Coarsened S2N (CoS2N) graph, formally,
\begin{equation}
\begin{aligned}
\sS^{\co} = \text{Coarsening}( \gG ),\ \ \hmA_{[i, j]}^{\co} = T_e (\gS_i, \gS_j)\\
\where\ (\gS_i, \gS_j) \in (\sS \cup \sS^{\co}) \times (\sS \cup \sS^{\co}).
\end{aligned}
\end{equation}
Here, any algorithm that coarsens the graph can be used for Coarsening($\gG$) (See \S\ref{sec:related_work}).

Training of CoS2N is done similarly to semi-supervised node classification. The virtual (unlabeled) subgraphs act as bridges to pass messages between labeled subgraphs. These allow S2N to better approximate the global graph that the existing set of subgraphs does not cover. We also show that adding virtual subgraphs to S2N can reduce the approximation error between representations of S2N and the global graph (Proposition~\ref{proposition:single_gcn_approx}).

The graph coarsening does not impair the efficiency for two reasons. First, it is performed only once before the training. Second, we can create a small CoS2N graph by tuning coarsening methods and their hyperparameters (e.g., the coarsening ratio).

\section{Theoretical Analysis on S2N}\label{sec:theoretical_analysis}

This section analytically compares the efficiency and the representation quality between S2N and the original graph. We first show how much S2N reduces computational complexity. Then, we analyze the error bound of representations between S2N and the original graph. All proofs are provided in Appendix~\ref{appendix:proof}.

\subsection{How Much Does S2N Reduce Computational Complexity?}

We introduce more notations for this analysis. For the global graph $\gG$ and the S2N graph $\hgG$, the numbers of edges are $E$ and $\hE$. Across a set $\sS$ of subgraphs, the average numbers of nodes and edges are $\overline{N^{\sub}}$ and $\overline{E^{\sub}}$. Note that $N$ is the number of nodes in $\gG$ and $M$ is the number of subgraphs.

In Proposition~\ref{proposition:time_complexity}, we compare the time complexity of single-layer GLASS (the state-of-the-art model)~\citep{wang2022glass}, Connected form, S2N+0, and S2N+A.

\propositionTimeComplexity \vspace{-0.1cm}
Considering that $N \ll E$ in real-world graphs~\citep{chung2010graph}, the significant difference between baselines and S2N is that $E$ becomes $\hE$. We can know that $\hE$ cannot be higher than $M^2$. The smaller $\overline{N^\sub}$, the smaller $\hE$ since it lowers the number of possible connections between nodes in subgraphs. However, it is difficult to directly compare $\hE$ and $E$ without assumptions of the global graph and subgraphs.

Thus, we investigate what $\hE$ takes in the random graph model as a global graph, specifically, the Configuration Model (CM). The CM graph of $N$ nodes is randomly generated from a given degree sequence $[d_1, d_2, ..., d_N]$~\citep{newman2018networks}. Note that CM can model graphs of arbitrary degree distributions, so its assumptions are mild compared to other models (See Appendix~\ref{appendix:cm} for details). For the CM global graph with independent and identically distributed (i.i.d.) subgraphs, we analyze the distribution of S2N's edge weights (i.e., $\hmA$) and demonstrate the probability that an edge weight is higher than a certain value. This probability is proportional to the number of S2N edges after normalization (Equation~\ref{eq:s2n_edge_normalization}). The smaller this is, the smaller $\hE$.
\propositionEdgeCM \vspace{-0.1cm}
It is well-known that degrees follow a power-law distribution in many real-world graphs~\citep{barabasi1999emergence}. Most nodes have a low degree; thus, the average degree $\E[d]$ has a small value. Proposition~\ref{proposition:edge_cm} implies that edges with small weights are more likely to appear in S2N, and the edge normalization can make the S2N graph sparse (i.e., small $\hE$). We also empirically confirm that edges in S2N are fewer than those of the global graph in Table~\ref{tab:s2n_statistics}.

\subsection{How Does S2N Approximate Subgraph Representations?}

For this subsection, we define the mapping matrix $\mM \in \{ 0, 1 \}^{N \times M}$, where $\mM_{[v, i]}$ is $1$ if and only if the node $v$ belongs to the subgraph $\gS_i$ (i.e., $\hmA = \mM^{\top} \mA \mM$). Degree matrices of $\gG$ and $\hgG$ are $\mD = \diag ( d_1, d_2, ..., d_N )$ and $\hmD = \diag ( \hd_1, \hd_2, ..., \hd_M )$. Also, $\lVert \cdot \rVert$ is the Frobenius norm.

This analysis aims to analytically compare node representations $\hmH \in \sR^{M \times F}$ of the S2N graph $\hgG$ and subgraph representations of the global graph $\gG$. Since outputs of GNN with the global graph are original nodes' representations $\mH \in \sR^{N \times F}$, we apply the readout to pool nodes in the subgraph where $\mR \in \sR^{N \times M}$ is a readout matrix:
\vspace{-0.13cm}
\begin{equation}\label{eq:readout_function}
\readout(\mH) = \mR^{\top} \mH \in \sR^{M \times F}.
\end{equation}
In this paper, we adopt a degree-dependent readout matrix $\mR$ inspired by configuration-based reconstruction~\citep{zhou2021dpgs, zhou2022learning}, which is defined as follows:
\vspace{-0.13cm}
\begin{equation}\label{eq:readout_matrix}
\mR = \mD^{\frac{1}{2}} \mM \hmD^{-\frac{1}{2}}
\quad
\text{i.e.,}\ \  
\mR_{[v, i]} = ( d_v / \hd_i )^{\frac{1}{2}}.
\end{equation}
We now demonstrate that the S2N's node representations $\hmH$ approximate the global graph's subgraph representations $\mR^{\top} \mH$, particularly when the model is a single-layer GCN. The error bound between $\hmH$ and $\mR^{\top} \mH$ is introduced in Proposition~\ref{proposition:single_gcn_approx}. We also conduct a similar analysis for Graph Isomorphism Networks~\citep{xu2018how} in Corollary~\ref{corollary:simple_single_gins1l_approx} (Appendix~\ref{appendix:proof}).

\propositionSingleGCNApprox \vspace{-0.1cm}
The error between representations is bounded by the error between input features $\mX$ and $\mR \hmX$. As in \citet{zhou2022learning}, if we use the initial features $\hmX = \mR^\top \mX$ for S2N, $\mR \hmX$ is $(\mR \mR^\top) \mX$. The matrix $\mR \mR^\top \in \sR^{N \times N}$ has rank $M (<N)$, then $\mR \hmX$ is a low-rank approximation of $\mX$. Since $\mR$ is given by subgraphs, $\mR \hmX$ may not sufficiently approximate $\mX$ for the downstream task. In particular, when there are only a few subgraph samples (i.e., very small rank $M$), the expressiveness of S2N can be weakened. This theoretical observation implies that the proposed CoS2N (\S\ref{sec:coarsening}) higher-rank approximates $\mX$ for a data-scarce setting.

\section{Experiments}\label{sec:experiments}

This section describes the experimental setup, including datasets, training, evaluation, and models.

\paragraph{Datasets}

We use four real-world datasets (\PPIBP, \HPONeuro, \HPOMetab, and \EMUser) and four synthetic datasets (\Density, \CutRatio, \Coreness, and \Component) introduced in \citet{alsentzer2020subgraph}. The task is subgraph classification, where the global graph $\gG$ and subgraphs $\sS$ are given. The input node features $\mX$ are pre-trained embedding from \citet{wang2022glass} for real-world datasets and constant features for synthetic datasets. Dataset statistics and descriptions are in Tables~\ref{tab:dataset_original}, \ref{tab:dataset_original_syn}, and Appendix~\ref{appendix:dataset}.

\paragraph{Training and evaluation}\label{para:t_and_e}

In the original setting from \citet{alsentzer2020subgraph}, evaluation (i.e., validation and test) subgraphs cannot be seen during the training stage. Following this protocol, we create different S2N graphs for each stage using train and evaluation sets of subgraphs ($\sS_{\txtrain}$ and $\sS_{\txeval}$). For the S2N translation, we use $\sS_{\txtrain}$ only in the training stage and use both $\sS_{\txtrain} \cup \sS_{\txeval}$ in the evaluation stage. We predict unseen nodes based on structures translated from $\sS_{\txtrain} \cup \sS_{\txeval}$ in the evaluation stage. In this respect, node classification on S2N-translated graphs is inductive.

\paragraph{Models}\label{para:models_for_s2n}

We use two well-known GNNs for $\gnn_{\STON}$: GCN~\citep{kipf2017semi} and GCNII~\citep{chen2020simple}. For the node-level encoder $\gnn_{\node}$ in S2N+A, we use the same kind of GNN as $\gnn_{\STON}$. See Appendix~\ref{appendix:model} for their hyperparameters. We also test these models for connected and separated forms.

\paragraph{Baselines}

We use basic and state-of-the-art models for subgraph classification tasks as baselines: Sub2Vec~\citep{adhikari2018sub2vec}, GBDT, SubGNN~\citep{alsentzer2020subgraph}, and GLASS~\citep{wang2022glass}. We report the best performance among the variants of Sub2Vec. All baseline results are reprinted from \citet{alsentzer2020subgraph} and \citet{wang2022glass}.

\paragraph{Efficiency measurement}\label{para:efficiency}

We use each model's best hyperparameters (including batch sizes) and take the mean wall-clock time over 50 epochs. Throughput and latency are all measured using training and validation sets for each stage. We count the number of all trainable parameters, including node embeddings. The maximum allocated GPU VRAM is measured by the PyTorch API. We fix the computation device as Intel(R) Xeon(R) CPU E5-2640 v4 and a single GeForce GTX 1080 Ti in measuring efficiency metrics. We describe details in Appendix~\ref{appendix:efficiency}.

\paragraph{Data-scarce experiments}

Experiments in a data-scarce setting are conducted on benchmarks with the smallest and largest global graphs (\PPIBPb and \EMUser), and we set the number of training samples per class to 10, 20, 40, and 80. To coarse the global graph, we employ the Variation Edges method~\citep{loukas2019graph} and select the coarsening ratio that generates subgraphs smaller than average sizes. All experiments use GCNII, which performs well across datasets in a fully supervised setting.

\section{Results and Discussions}\label{sec:results}


\begin{table}[t]
\centering
\caption{The number of nodes and edges of real-world graphs before and after S2N translation. The number of edges in S2N is averaged across all experiments.}
\label{tab:s2n_statistics}
\resizebox{\columnwidth}{!}{%
\begin{tabular}{llllll}
\hline
        &          & \PPIBP              & \HPONeuro           & \HPOMetab           & \EMUser             \\ \hline
\# Nodes & Original & $ 1.7 \times 10^4 $ & $ 1.5 \times 10^4 $ & $ 1.5 \times 10^4 $ & $ 5.7 \times 10^4 $ \\
        & S2N      & $ 1.6 \times 10^3 $ & $ 4.0 \times 10^3 $ & $ 2.4 \times 10^3 $ & $ 3.2 \times 10^2 $ \\ \hline
\# Edges & Original & $ 3.2 \times 10^5 $ & $ 3.2 \times 10^6 $ & $ 3.2 \times 10^6 $ & $ 4.6 \times 10^6 $ \\
        & S2N  & $ 4.8 \times 10^3 $ & $ 6.3 \times 10^5 $ & $ 6.0 \times 10^4 $ & $ 3.0 \times 10^3 $ \\ \hline
\end{tabular}%
}
\end{table} 
\begin{table}[]
\centering
\caption{Mean performance in micro F1-score on real-world datasets over 10 runs. For the top 50\% of results, the higher the performance, the darker the blue color. The unpaired \ttest\ result with the best is denoted by superscripts at a level of 0.01 ($*$: significantly outperforms, $\sim$: no significant difference). We mark with daggers reprinted results from \citet{alsentzer2020subgraph} ($\dagger$) and \citet{wang2022glass} ($\ddagger$).}
\label{tab:results_real}
\resizebox{\columnwidth}{!}{%
\begin{tabular}{lllll}
\hline
Model                   & \PPIBP                                          & \HPONeuro                                       & \HPOMetab                                       & \EMUser                                         \\ \hline
Sub2Vec$^\dagger$       & \cellcolor[HTML]{FFFFFF}$30.9_{\pm 2.3}$        & \cellcolor[HTML]{FFFFFF}$22.3_{\pm 6.5}$        & \cellcolor[HTML]{FFFFFF}$13.2_{\pm 4.7}$        & \cellcolor[HTML]{FFFFFF}$85.9_{\pm 1.4}$        \\
GBDT$^\ddagger$         & \cellcolor[HTML]{FFFFFF}$44.6_{\pm 0.0}$        & \cellcolor[HTML]{FFFFFF}$51.3_{\pm 0.0}$        & \cellcolor[HTML]{FFFFFF}$40.4_{\pm 0.0}$        & \cellcolor[HTML]{FFFFFF}$69.4_{\pm 0.0}$        \\
SubGNN$^\dagger$        & \cellcolor[HTML]{FFFFFF}$59.9_{\pm 2.4}$        & \cellcolor[HTML]{FFFFFF}$63.2_{\pm 1.0}$        & \cellcolor[HTML]{FFFFFF}$53.7_{\pm 2.3}$        & \cellcolor[HTML]{FFFFFF}$81.4_{\pm 4.6}$        \\
GLASS$^\ddagger$        & \cellcolor[HTML]{FFFFFF}$61.9_{\pm 0.7}$        & \cellcolor[HTML]{A9C6F5}$68.5_{\pm 0.5}$        & \cellcolor[HTML]{FFFFFF}$61.4_{\pm 0.5}$        & \cellcolor[HTML]{AAC6F5}$88.8_{\pm 0.6}$        \\ \hline
Sep.$_{\text{ GCN}}$    & \cellcolor[HTML]{FFFFFF}$61.4_{\pm 2.0}$        & \cellcolor[HTML]{D4E3FA}$67.6_{\pm 1.0}$        & \cellcolor[HTML]{FFFFFF}$60.1_{\pm 2.8}$        & \cellcolor[HTML]{FFFFFF}$84.5_{\pm 4.1}$        \\
Sep.$_{\text{ GCNII}}$  & \cellcolor[HTML]{FFFFFF}$61.3_{\pm 1.2}$        & \cellcolor[HTML]{D0DFFA}$67.7_{\pm 0.6}$        & \cellcolor[HTML]{FFFFFF}$59.4_{\pm 2.7}$        & \cellcolor[HTML]{FFFFFF}$84.7_{\pm 4.1}$        \\
Con.$_{\text{ GCN}}$    & \cellcolor[HTML]{FFFFFF}$62.6_{\pm 1.7}$        & \cellcolor[HTML]{FFFFFF}$65.7_{\pm 0.8}$        & \cellcolor[HTML]{FFFFFF}$60.6_{\pm 2.0}$        & \cellcolor[HTML]{FFFFFF}$85.9_{\pm 2.8}$        \\
Con.$_{\text{ GCNII}}$  & \cellcolor[HTML]{DFEAFC}$63.5_{\pm 2.0}$        & \cellcolor[HTML]{FFFFFF}$66.7_{\pm 0.8}$        & \cellcolor[HTML]{FDFEFF}$61.7_{\pm 2.7}$        & \cellcolor[HTML]{FFFFFF}$85.5_{\pm 4.8}$        \\ \hline
S2N+0$_{\text{ GCN}}$   & \cellcolor[HTML]{FFFFFF}$63.0_{\pm 2.3}^{\sim}$ & \cellcolor[HTML]{FFFFFF}$66.4_{\pm 0.7}$        & \cellcolor[HTML]{F1F6FE}$62.0_{\pm 1.6}^{\sim}$ & \cellcolor[HTML]{FFFFFF}$85.7_{\pm 2.9}$        \\
S2N+0$_{\text{ GCNII}}$ & \cellcolor[HTML]{DFEAFC}$63.5_{\pm 2.4}^{\sim}$ & \cellcolor[HTML]{FFFFFF}$66.4_{\pm 1.1}$        & \cellcolor[HTML]{FFFFFF}$61.6_{\pm 1.7}^{\sim}$ & \cellcolor[HTML]{EEF4FD}$86.5_{\pm 3.2}^{\sim}$ \\
S2N+A$_{\text{ GCN}}$   & \cellcolor[HTML]{EDF3FD}$63.3_{\pm 2.3}^{\sim}$ & \cellcolor[HTML]{B3CCF6}$68.3_{\pm 0.9}^{\sim}$ & \cellcolor[HTML]{F1F6FE}$62.0_{\pm 3.0}^{\sim}$ & \cellcolor[HTML]{EEF4FD}$86.5_{\pm 2.3}$        \\
S2N+A$_{\text{ GCNII}}$ & \cellcolor[HTML]{D0E0FA}$63.7_{\pm 2.3}^{\sim}$ & \cellcolor[HTML]{AEC9F6}$68.4_{\pm 1.0}^{\sim}$ & \cellcolor[HTML]{C1D5F8}$63.2_{\pm 2.7}^{\sim}$ & \cellcolor[HTML]{A4C2F4}$89.0_{\pm 1.6}^{\sim}$ \\ \hline
\textit{with RWPE}                &                                                 &                                                 &                                                 &                                                 \\
S2N+0$_{\text{ GCN}}$   & \cellcolor[HTML]{FCFDFF}$63.1_{\pm 2.2}^{\sim}$ & \cellcolor[HTML]{FFFFFF}$66.7_{\pm 0.6}$        & \cellcolor[HTML]{E5EEFC}$62.3_{\pm 1.9}^{\sim}$ & \cellcolor[HTML]{FFFFFF}$85.9_{\pm 2.8}$        \\
S2N+0$_{\text{ GCNII}}$ & \cellcolor[HTML]{DFEAFC}$63.5_{\pm 1.7}^{\sim}$ & \cellcolor[HTML]{FFFFFF}$66.7_{\pm 0.6}$        & \cellcolor[HTML]{E5EEFC}$62.3_{\pm 1.1}^{\sim}$ & \cellcolor[HTML]{EEF4FD}$86.5_{\pm 4.7}^{\sim}$ \\
S2N+A$_{\text{ GCN}}$   & \cellcolor[HTML]{E6EEFC}$63.4_{\pm 2.0}^{\sim}$ & \cellcolor[HTML]{AEC9F6}$68.4_{\pm 0.7}^{\sim}$ & \cellcolor[HTML]{F5F9FE}$61.9_{\pm 2.3}^{\sim}$ & \cellcolor[HTML]{D6E4FB}$87.3_{\pm 9.7}^{\sim}$ \\
S2N+A$_{\text{ GCNII}}$ & \cellcolor[HTML]{A4C2F4}$64.3_{\pm 1.8}^{*}$    & \cellcolor[HTML]{A4C2F4}$68.6_{\pm 0.8}^{\sim}$ & \cellcolor[HTML]{A4C2F4}$63.9_{\pm 1.7}^{*}$    & \cellcolor[HTML]{A4C2F4}$89.0_{\pm 3.1}^{\sim}$ \\ \hline
\end{tabular}%
}
\end{table}
\begin{table}[t]
\centering
\caption{Mean performance in micro F1-score on synthetic datasets over 10 runs. For the top 50\% of results, the higher the performance, the darker the blue color. The unpaired \ttest\ result between S2N and the best is denoted by superscripts ($\sim$: no significant difference at a level of 0.01). We mark with daggers the reprinted results from \citet{alsentzer2020subgraph} ($\dagger$) and \citet{wang2022glass} ($\ddagger$).}
\label{tab:results_syn}
\vspace{-0.1cm}
\resizebox{\columnwidth}{!}{%
\begin{tabular}{lllll}
\hline
Model                   & \Density                                        & \CutRatio                                & \Coreness                                       & \Component                                       \\ \hline
Sub2Vec$^\ddagger$      & \cellcolor[HTML]{FFFFFF}$45.9_{\pm 1.2}$        & \cellcolor[HTML]{FFFFFF}$35.4_{\pm 1.4}$ & \cellcolor[HTML]{FFFFFF}$36.0_{\pm 1.9}$        & \cellcolor[HTML]{FFFFFF}$65.7_{\pm 1.7}$         \\
SubGNN$^\dagger$        & \cellcolor[HTML]{FFFFFF}$91.9_{\pm 1.6}$        & \cellcolor[HTML]{FFFFFF}$62.9_{\pm 3.9}$ & \cellcolor[HTML]{FFFFFF}$65.9_{\pm 9.2}$        & \cellcolor[HTML]{FFFFFF}$95.8_{\pm 9.8}$         \\
GLASS$^\ddagger$        & \cellcolor[HTML]{C5D8F8}$93.0_{\pm 0.9}$        & \cellcolor[HTML]{A4C2F4}$93.5_{\pm 0.6}$ & \cellcolor[HTML]{ACC8F5}$84.0_{\pm 0.9}$        & \cellcolor[HTML]{A4C2F4}$100.0_{\pm 0.0}$        \\ \hline
S2N+0$_{\text{ GCNII}}$ & \cellcolor[HTML]{FFFFFF}$67.2_{\pm 2.4}$        & \cellcolor[HTML]{FFFFFF}$56.0_{\pm 0.0}$ & \cellcolor[HTML]{FFFFFF}$57.0_{\pm 4.9}$        & \cellcolor[HTML]{A4C2F4}$100.0_{\pm 0.0}^{\sim}$ \\
S2N+A$_{\text{ GCNII}}$ & \cellcolor[HTML]{BAD1F7}$93.2_{\pm 2.6}^{\sim}$ & \cellcolor[HTML]{FFFFFF}$56.0_{\pm 0.0}$ & \cellcolor[HTML]{A4C2F4}$85.7_{\pm 5.8}^{\sim}$ & \cellcolor[HTML]{A4C2F4}$100.0_{\pm 0.0}^{\sim}$ \\ \hline
\textit{with RWPE}                &                                                 &                                                 &                                                 &                                                 \\
S2N+0$_{\text{ GCNII}}$    & \cellcolor[HTML]{FFFFFF}$74.8_{\pm 3.6}$        & \cellcolor[HTML]{BDD3F7}$85.2_{\pm 5.1}$ & \cellcolor[HTML]{FFFFFF}$56.1_{\pm 3.0}$        & \cellcolor[HTML]{A4C2F4}$100.0_{\pm 0.0}^{\sim}$ \\
S2N+A$_{\text{ GCNII}}$  & \cellcolor[HTML]{A4C2F4}$93.6_{\pm 2.0}^{\sim}$ & \cellcolor[HTML]{B1CBF6}$89.2_{\pm 2.6}$ & \cellcolor[HTML]{E1EBFC}$72.6_{\pm 6.2}$        & \cellcolor[HTML]{A4C2F4}$100.0_{\pm 0.0}^{\sim}$ \\ \hline
\end{tabular}%
}
\end{table}

We analyze the characteristics of S2N graphs and compare our models and baselines on classification performance and efficiency. We show that S2N translation results in graph compression (\S\ref{sec:result_analysis_of_s2n}), which results in better or similar classification accuracy (\S\ref{sec:result_performance}) but significantly improves efficiency in terms of computation and memory (\S\ref{sec:result_efficiency}). Finally, we study Coarsened S2N (CoS2N)'s performance and efficiency in a data-scare setting (\S\ref{sec:result_efficiency_by_num_training}).

\subsection{Analysis of S2N-Translated Graphs}\label{sec:result_analysis_of_s2n}

Table~\ref{tab:s2n_statistics} summarizes the number of nodes and edges before and after S2N translation. These statistics are from S2N graphs (S2N+0 and S2N+A) tuned for the best performance on GCN and GCNII. The translated graphs have a smaller number of nodes ($\times 0.006$ -- $\times 0.27$) and edges ($\times 10^{-4}$ -- $\times 0.45$) than the original graphs. See Appendix~\ref{appendix:number_of_nodes} for detailed discussion. We also find that they are non-homophilous, meaning many connected node pairs differ in their class. The edge homophily of S2N graphs is $0.25 \pm 0.01$ for \PPIBP, $0.20 \pm 0.03$ for \HPONeuro\footnote{We propose multi-label edge homophily for multi-label datasets (\sHPONeuro). It generalizes the existing multi-class homophily, and we discuss more in Appendix~\ref{appendix:homophily_ml}.}, $0.24 \pm 0.01$ for \HPOMetab, and $0.51 \pm 0.01$ for \EMUser.

\subsection{Performance}\label{sec:result_performance}

\paragraph{Real-World Datasets}

In Table~\ref{tab:results_real}, we report the micro F1-score on real-world datasets. We confirm that S2N with simple GNN models outperforms or is similar to GLASS, the state-of-the-art model. In 16 experiments (4 datasets and 4 models), S2N models outperform GLASS in 9 cases and SubGNN in all 16 cases. Moreover, S2N models are on par with GLASS in 12 of 16 experiments; they have no significant difference at the level of 0.01. The best models with S2N show 102.9\% (\PPIBP), 99.9\% (\HPONeuro), 102.9\% (\HPOMetab), and 100.2\% (\EMUser) of GLASS's performances. We interpret that message-passing between subgraphs in S2N improves performance by capturing distant interactions that cannot occur in message-passing between nodes in the global graph. We also observe that RWPE generally increases performance. S2N models with RWPE are on par with GLASS in 14 cases and outperform in 13 cases. Plus, S2N+A outperforms S2N+0; internal structure also contributes to subgraph representation. However, the importance of internal structures varies across datasets. For \HPONeurob as an example, the performance improvement of S2N+A over S2N+0 is high compared to other datasets.

\paragraph{Synthetic Datasets}

In Table~\ref{tab:results_syn}, we summarize the performance of S2N models with GCNII and baselines on synthetic datasets. S2N+A performs better than or the same as the state-of-the-art (GLASS) on \Density, \Coreness, and \Component. S2N+0 shows the same performance as GLASS only in \Component. We can explain these results through known subgraph properties associated with synthetic labels (Table~\ref{tab:syn_properties}). Because S2N compresses the global graph structure, it is challenging to learn \CutRatio, which requires exact information about the global structure. Learning the density and coreness of subgraphs requires their internal structures. Therefore, S2N+0, which does not maintain internal structure, relatively underperforms baselines.
RWPE allows S2N to improve the performance of \Densityb and \CutRatiob significantly, but not of \Coreness. We interpret that RWPE for subgraphs can encode internal and border structures well but cannot encode border positions. We leave the development of structural encoding for S2N as future work.

\subsection{Efficiency}\label{sec:result_efficiency}

\begin{figure*}[t]
  \centering
  \begin{subfigure}[t]{\textwidth}
    \centering
    \includegraphics[width=\textwidth]{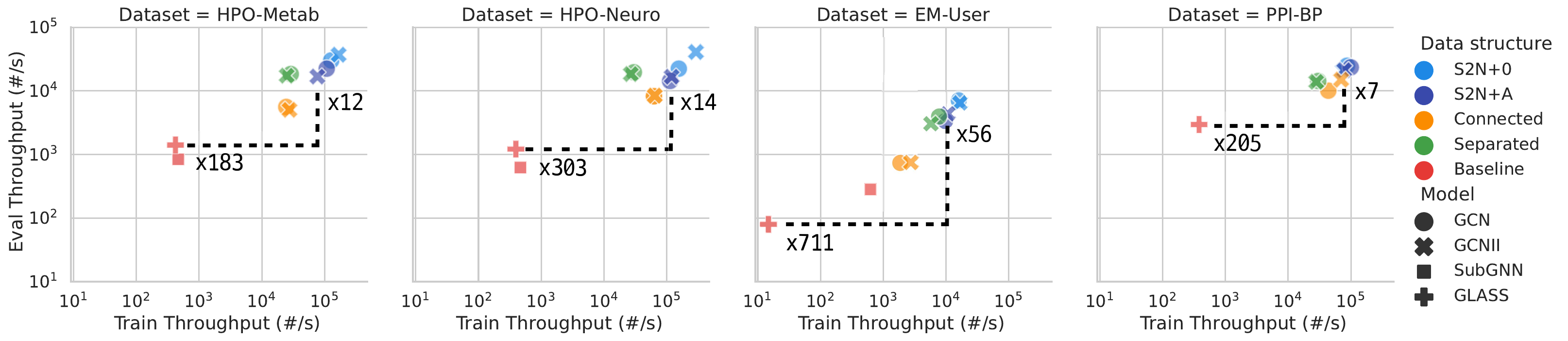}
    \caption{The throughput (the number of subgraphs / second) at training and evaluation stages. \textit{The higher, the better.}}
    \vspace{0.2cm}
    \label{fig:efficiency_throughput}
  \end{subfigure}
  \begin{subfigure}[t]{\textwidth}
    \centering
    \includegraphics[width=\textwidth]{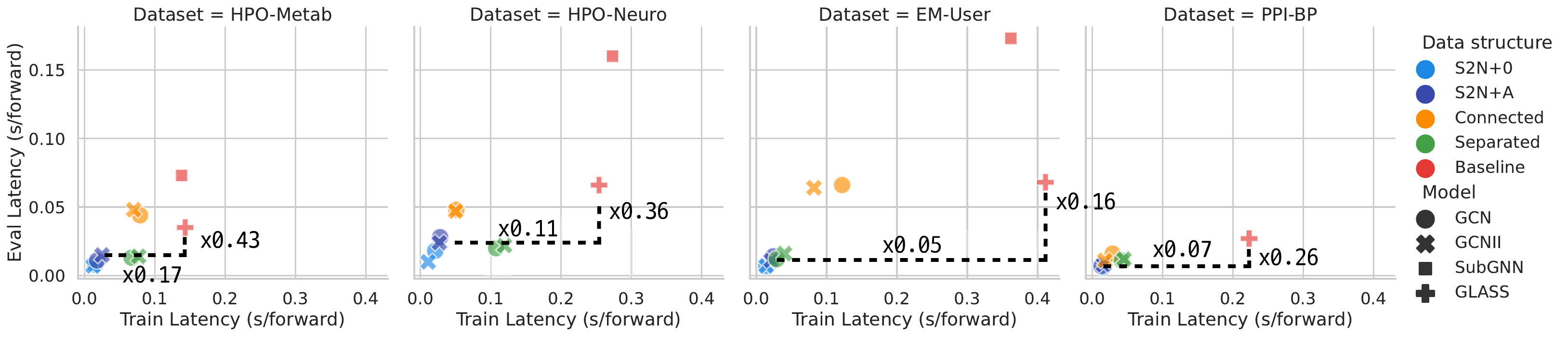}
    \caption{The latency (seconds / forward pass) at training and evaluation stages. \textit{The lower, the better.}}
    \vspace{0.2cm}
    \label{fig:efficiency_latency}
  \end{subfigure}
  \begin{subfigure}[t]{\textwidth}
    \centering
    \includegraphics[width=\textwidth]{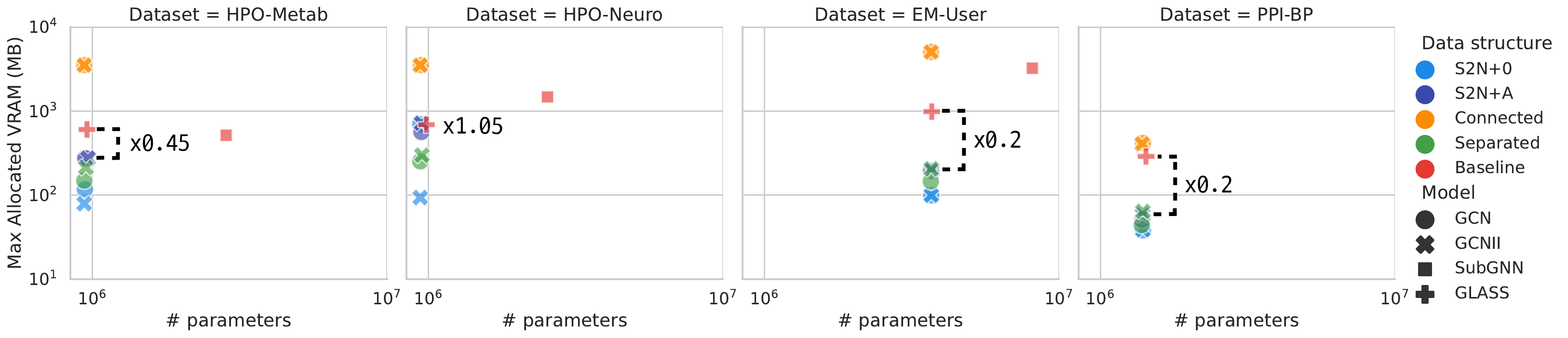}
    \caption{The number of parameters and maximum allocated GPU VRAM. \textit{The lower, the better.}}
    \label{fig:efficiency_num_param_vram}
  \end{subfigure}
  \vspace{-0.1cm} 
  \caption{Efficiency of S2N models and baselines on real-world datasets. The ratio of the best S2N model and the state-of-the-art model for each metric is notated in the figure (dashed lines).}
  \label{fig:efficiency_all}
\end{figure*}

\begin{figure*}[t]
  \centering
  \hspace{-1.4cm}
  \begin{subfigure}[b]{0.24\textwidth}
    \centering
    \includegraphics[height=0.145\textheight]{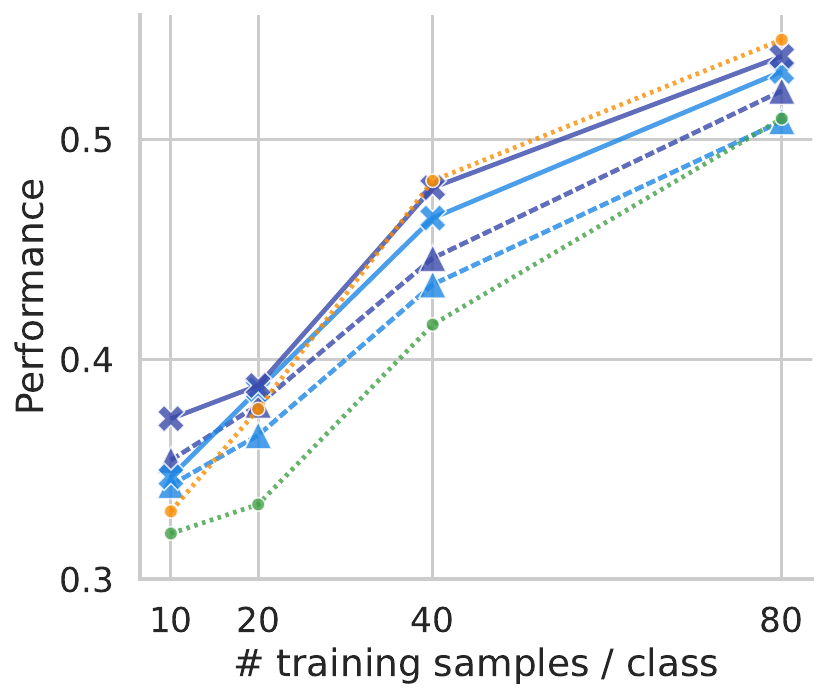}
    \caption{Performance}
    \label{fig:performance_by_num_training}
  \end{subfigure}
  \hspace{-0.35cm}
  \begin{subfigure}[b]{0.24\textwidth}
    \centering
    \includegraphics[height=0.145\textheight]{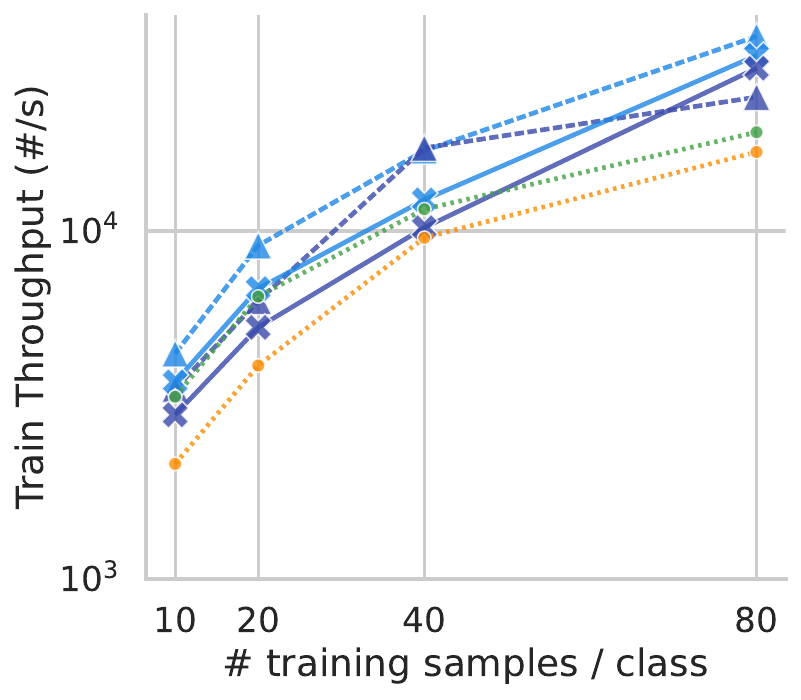}
    \caption{Training Throughput}
    \label{fig:training_throughput_by_num_training}
  \end{subfigure}
  \hspace{-0.35cm}
  \begin{subfigure}[b]{0.24\textwidth}
    \centering
    \includegraphics[height=0.145\textheight]{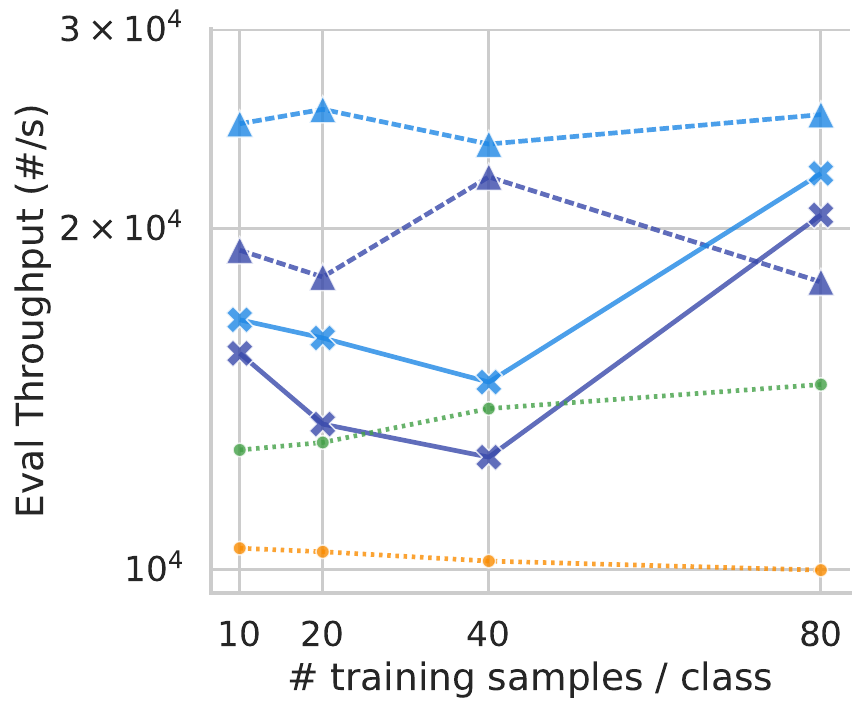}
    \caption{Eval Throughput}
    \label{fig:evaluation_throughput_by_num_training}
  \end{subfigure}
  \begin{subfigure}[b]{0.24\textwidth}
    \centering
    \includegraphics[height=0.145\textheight]{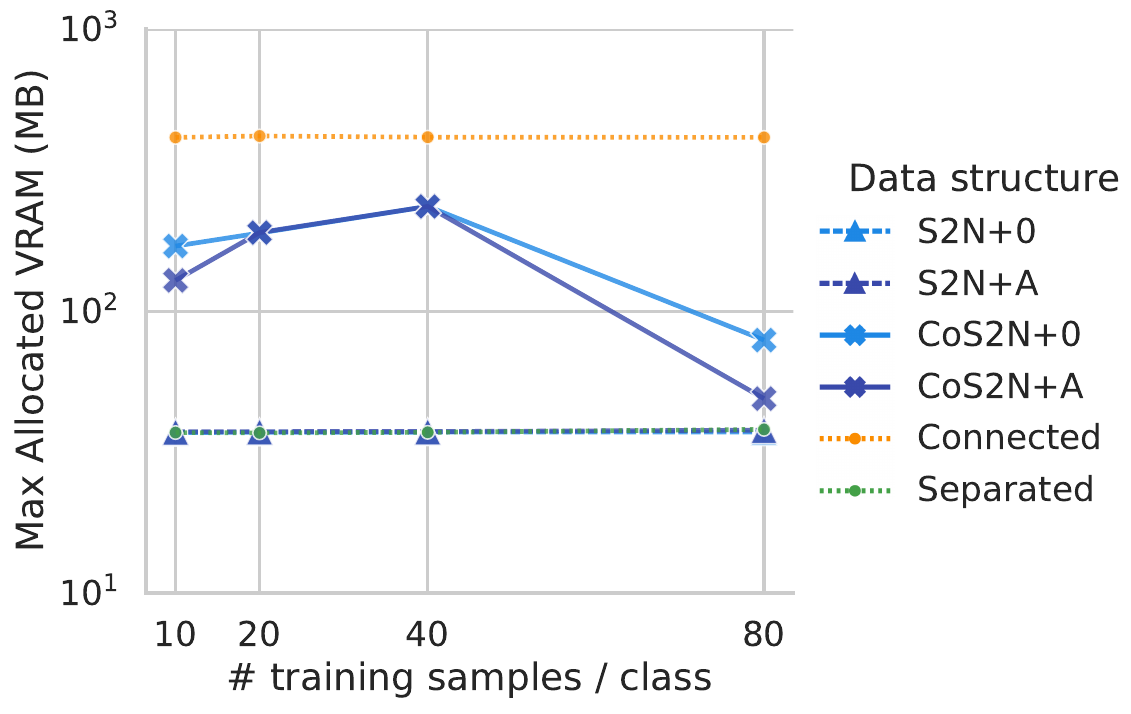}
    \caption{Max allocated VRAM}
    \label{fig:gpu_vram_by_num_training}
  \end{subfigure}
  \caption{Performance and efficiency on \PPIBPb of S2N, CoS2N, connected, and separated forms by the number of training samples in a data-scarce setting.}
  \label{fig:efficiency_by_num_training}
\end{figure*}

In Figure~\ref{fig:efficiency_all}, we show throughput, latency, the number of parameters, and the maximum allocated GPU VRAM of two models with three data structures and state-of-the-art baselines. We cannot experiment on \PPIBPb with SubGNN since it takes more than 48 hours in pre-computation. We make the following five observations from these results.

\paragraph{S2N models show significantly high throughput (Fig.~\ref{fig:efficiency_throughput}).}

The best S2N models can process $\times 183$ -- $\times 711$ more samples than the state-of-the-art model (GLASS) for the same training time. At the evaluation stage, they show $7$ -- $56$ times higher throughput than GLASS. This difference is not as large as the training stage, but S2N is still significantly more efficient than GLASS. In addition, S2N shows higher training throughput than connected and separated forms. 

\paragraph{S2N models even with full batch show lower latency than others with small batch size (Fig.~\ref{fig:efficiency_latency}).}

Comparing the best S2N model and GLASS, the training latency is $\times0.05$ -- $\times0.17$ and evaluation latency is $\times0.16$ -- $\times0.43$. Note that measuring latency ignores the parallelism from large batch sizes. S2N's superiority over other data structures can be underestimated in latency rather than throughput because it requires full batch computation. Note that existing models need to use small batch sizes because of intensive memory requirements (SubGNN) or model design (GLASS).

\paragraph{S2N models require less memory even with a similar level of parameters (Fig.~\ref{fig:efficiency_num_param_vram}).}

For a given dataset, the number of parameters of each model does not vary much, but the GPU VRAM in actual runtime varies by a large margin. The best models with S2N need less memory ($\times0.2$ -- $\times0.45$) than GLASS except for \HPONeuro. \HPONeuro, which has a large number of subgraphs, requires the same level of memory ($\times1.05$). In particular, since S2N does not employ a large global graph, S2N works with only $\times 0.13$ memory on average compared to the connected form.

\paragraph{S2N+A does not show a significant difference from S2N+0 in efficiency.}

Recall that S2N+A differs from S2N+0 by using the internal edges of subgraphs. However, the number of internal edges is negligible compared to the original global edges, as in Table~\ref{tab:dataset_original}. Consequently, the added internal edges require only a small amount of additional computation and memory, allowing S2N+A to perform training and inference efficiently.

\paragraph{Overall, S2N models outperform baselines in all computational and memory efficiency metrics.}

Models with S2N process many samples faster (i.e., higher throughput and lower latency), and require less GPU memory than other data structures and state-of-the-art models. The separated form, which does not use a global graph, shows a similar level of efficiency as S2N in some experiments but loses performance by completely ignoring the global structure.

\subsection{Performance and Efficiency in a Data-Scarce Setting}\label{sec:result_efficiency_by_num_training}

In this section, we report the performance and efficiency of S2N, Coarsened S2N (CoS2N), connected form, and separated form. Figure~\ref{fig:efficiency_by_num_training} summarizes the performance, throughput, and max allocated VRAM by the number of training samples on \PPIBP. In Appendix~\ref{appendix:efficiency_all_by_num_training}, we analyze similar results on the other dataset and the ablation study on the coarsening ratio.

\paragraph{Virtual subgraphs created in Coarsened S2N contribute to performance improvements of S2N (Figure~\ref{fig:performance_by_num_training}).} 

We observe that CoS2N consistently outperforms S2N in all conditions. This implies that virtual subgraphs created from graph coarsening in CoS2N enhance communications between subgraphs, leading to better representations. CoS2N+A generally surpasses all other models, including CoS2N+0 and the connected form. When the number of training samples is extremely small, both S2N and CoS2N demonstrate superior performance compared to both baseline models. We confirm that the message-passing between subgraphs is more effective when supervised signals are scarce. In conclusion, CoS2N approximates representations of the global graph well, even though the virtual subgraphs created through coarsening do not follow the distribution of real subgraphs.

\paragraph{CoS2N has higher throughput (Figures~\ref{fig:training_throughput_by_num_training}, ~\ref{fig:evaluation_throughput_by_num_training}) and uses less memory (Figure~\ref{fig:gpu_vram_by_num_training}) than using the global graph.}

Although the virtual subgraphs by coarsening are added, both CoS2N methods show higher throughput than using the global graph (i.e., the connected form). CoS2N+0 even shows higher throughput than the separated form in all stages. CoS2N+A shows higher throughput than the separated form in the evaluation stage, where there are more subgraphs to be processed. The training throughput increases as more training samples are used since the full batch parallelization of GPUs can efficiently process additional samples.

Like computational requirements, CoS2N uses less memory than the connected form. This is because graph coarsening creates fewer subgraphs than the size of the global graph. By the training set size, the memory consumption is constant for the connected form and fluctuates for CoS2N. The memory bottleneck of the connected form and CoS2N is the largest component of each dataset: the global graph and coarsened nodes, respectively. Adding training samples does not substantially affect memory demand; instead, the size of the coarse graph does for CoS2N.

\section{Limitations and Future Work}\label{sec:limitations}

This section discusses the limitations and future work. First, a large S2N graph will be generated if there are excessive subgraphs. This is because the current method translates all subgraphs into nodes in the S2N graph. Future research is needed to sample important subgraphs that sparsely summarize the original global graph. Second, we can observe that the performance of S2N varies depending on the dataset, but this can be known once experiments are conducted. Practitioners can only know which data structures or models are appropriate for a given task by empirical experiments. We leave it to future work to identify which subgraph characteristics affect the performance of models, including S2N. Lastly, we need synthetic tasks for better evaluation. For synthetic datasets, labels are created by pre-designed rules that depend only on structures (e.g., density, cut ratio, and core numbers). These datasets reflect only a very narrow aspect of a subgraph's properties. However, the labels of real-world subgraphs depend on various information about the structures and features. This implies a gap between synthetic and real-world subgraphs, and future studies are required to develop more realistic synthetic tasks.

\section{Conclusion}\label{sec:conclusion}
Subgraph-to-node (S2N) translation is a novel, efficient way to learn representations of subgraphs. S2N takes the original subgraphs and creates a new graph where the nodes are the subgraphs, and the edges are the relations between the subgraphs, thereby performing subgraph-level tasks as node-level tasks. We empirically and theoretically show that S2N translation significantly reduces memory and computation costs without performance degradation. Specifically, the best-performing models with S2N on real-world datasets show $\times 183 - \times 711$ of throughput and achieve at least 99.9\% of the state-of-the-art models for classification performance.

\section*{Acknowledgements}

This research was supported by the Engineering Research Center Program through the National Research Foundation of Korea (NRF) funded by the Korean Government MSIT (NRF-2018R1A5A1059921)

\section*{Impact Statement}

This paper aims to advance the field of graph representation learning, and it is difficult to expect direct societal consequences. Instead, we discuss a potential concern about bias and fairness that has not been confirmed in theory or practice. Models with S2N pass messages between all samples within the batch; thus, the majority in a batch can be over-represented during prediction and the bias in representations can be amplified. The tasks covered in this paper are unrelated to this, but practitioners should be aware of this for tasks that contain sensitive attributes.

\clearpage


\bibliography{example_paper}
\bibliographystyle{icml2024}

\newpage
\appendix
\onecolumn

\section{Reproducibility Statements}

To reproduce the results, we open our code public via the GitHub link\footnote{\href{https://github.com/dongkwan-kim/S2N}{\url{https://github.com/dongkwan-kim/S2N}}}. Datasets, including the downloadable link from~\citet{alsentzer2020subgraph}, are described in Appendix~\ref{appendix:dataset}.

\section{Discussion on Related Work}\label{appendix:related_work}

\subsection{Detailed Comparison with State-of-the-Art Models: SubGNN and GLASS}\label{appendix:architectural_diff}

SubGNN~\citep{alsentzer2020subgraph}, GLASS~\citep{wang2022glass}, and S2N improve different parts of the machine learning pipeline to solve subgraph-level tasks. SubGNN designs a whole model, GLASS augments input data through a labeling trick, and S2N uses a new data structure.

SubGNN performs message-passing between subgraphs (or patches). Through this, the properties of internal and border structures for three channels (position, neighborhood, and structure) are learned independently. To learn a total of 6 (2 × 3) properties, SubGNN designs patch samplers, patch representation, and similarity (weights of messages) for each property in an ad hoc manner. For example, SubGNN patches nodes inside the subgraph using its representation as a message and distance-based similarity as weights to learn internal positions. By the complex model design, SubGNN requires a lot of computational resources for data pre-processing, model training, and inference.

GLASS uses plain GNNs but labels input nodes as to whether they belong to the subgraph (the label of one) or not (the label of zero). Separate node-level message-passing is performed for each label to distinguish the internal and border structures of the subgraph. GLASS's labeling trick is effective, but handling multiple labels from multiple subgraphs in a batch is hard. Although the authors of GLASS propose a max-zero-one trick to address this issue, small batches are still recommended. In addition, using a large global graph requires significant computational and memory resources.

S2N uses the new data structure that stores and processes subgraphs efficiently. By compressing the global graph, computational and memory resource requirements are reduced. There are no restrictions on batch sizes; thus, we can train S2N graphs in the full batch. Our contribution lies in addressing the current research direction in subgraph representation learning, where enhancements in performance are often achieved by increasing model complexity. The S2N model challenges it by demonstrating better or comparable performance through a more efficient approach. This efficiency does not merely encompass computational resources. Still, it extends to ease of implementation and adaptability to diverse tasks, making it a significant advancement over current state-of-the-art methods.

\subsection{Detailed Comparison with Similar Architectures}\label{appendix:similar_architectures}

Circuit Graph Neural Network (CktGNN)~\citep{dong2022cktgnn} and Nested Graph Neural Network (NGNN)~\citep{zhang2021nested} are similar to our S2N+A in that they employ two-level GNNs, where the first GNN learns the embedding of subgraphs, and the second GNN performs message-passing between subgraphs. However, our study differs from these in the following aspects. First, they do not focus on subgraph-level tasks. CktGNN is applied for circuit design automation, and NGNN is applied for graph regression and classification. They do not demonstrate how the two-level GNNs approach affects subgraph representations rather than the whole graph. In contrast, we analyzed our S2N models empirically and theoretically on subgraph representation learning. Second, they also make strong assumptions about subgraphs and cannot be generalized to all subgraph-level tasks. Specifically, CktGNN uses a set of pre-designed subgraphs specialized for circuits. A subgraph in NGNN is the rooted subgraph of each node in the given graph. For these two models, the subgraph has to be in a fixed shape; thus, they cannot handle subgraphs of various structures and sizes.

Junction Tree Variational Autoencoder (JT-VAE)~\citep{jin2018junction} decomposes a molecular graph into a junction tree, where a node corresponds to the motif (particularly a ring of atoms), and edges link the nodes that share the nodes. This method is a graph generation model to learn the ring substructure well in chemical tasks but has not been used in subgraph-level tasks. Due to the nature of the Junction Tree algorithm, only the ring (or cycle) structure of input graphs is used as subgraphs, and the output is restricted to trees, which leads to limited usage. Our proposed S2N's primary contribution is to explore the fundamental question of subgraph representation learning and propose a novel perspective. In addition, S2N can be generally applied to graphs and subgraphs of any structure.

DiffPool~\citep{ying2018hierarchical} learns the hierarchy of a graph to obtain graph-level representations. DiffPool softly assigns each node to a cluster during training by optimizing the downstream task loss. To stabilize the soft clustering assignment, the authors of DiffPool employ link prediction loss and entropy regularization loss. The problem is that the assignment matrix must be maintained in GPU memory, which requires quadratic memory complexity regarding the number of nodes. In other words, we cannot apply DiffPool to large graphs such as global graphs in our use cases. We aim to perform subgraph representation learning efficiently by compressing data and reducing GPU load. Memory-intensive graph coarsening, such as DiffPool's soft clustering assignment, should not be used to keep CoS2N efficient. Instead, we can secure the efficiency of CoS2N by performing graph coarsening before training the model, relying only on the structure of the global graph.

\section{Justification for the Choice of the Random Graph Model}\label{appendix:cm}

The configuration model (CM) only requires a degree sequence or a distribution. That means CM can also generate graphs generated by other random graph models. For example, when the degree distribution is Poisson distribution, CM generates graphs close to the Erdős–Rényi model. CM can also adopt other degree distributions, for example, power-law distributions. See \citet{newman2018networks} for more details.

We also emphasize that the complexity of S2N strongly depends on the number of edges in S2N, that is, how many edges of small weights are removed during normalization. Thus, we need a random graph model that can analytically calculate the distribution of edge weights (i.e., the number of shared edges in two subgraphs). When using the CM, the distribution of S2N's edge weights can be derived from the degree distribution of the global graph. This is possible because CM calculates the probability of edge existence through the degrees of a pair of nodes. Note that CM is frequently used in analytically calculating numerous network measures~\citep{barabasi2013network}.

\section{Proofs of Theoretical Analysis}\label{appendix:proof}
This section describes proofs of theoretical claims in the paper: Propositions~\ref{proposition:time_complexity}, \ref{proposition:edge_cm}, and \ref{proposition:single_gcn_approx}. In addition, we analyze the error bound between S2N and the global graph for a variant of Graph Isomorphism Networks~\citep{xu2018how} in Proposition~\ref{proposition:single_gins1l_approx} and Corollary~\ref{corollary:simple_single_gins1l_approx}.

\vspace{0.2cm}

\begin{proposition}[Proposition~\ref{proposition:time_complexity}]
\propositionTimeComplexityContent
\end{proposition}
\begin{proof}
Let $\gG = (\sV, \mA, \mX)$ be a global graph where $\sV$ is a set of nodes ($|\sV| = N$), $\mA \in \{ 0, 1 \}^{N \times N}$ is an adjacency matrix, and $\mX \in \sR^{N \times F_0}$ is a node feature matrix. A subgraph $\gS = (\sV^{\sub}, \mA^{\sub})$ is a graph formed by subsets of nodes and edges in the global graph $\gG$. For the subgraph classification task, there is a set of $M$ subgraphs $\sS = \{ \gS_1, \gS_2, ..., \gS_M \}$, and for $\gS_i = (\sV_i^{\sub}, \mA_i^{\sub})$, the goal is to learn subgraph representations $\hmH \in \sR^{M \times F}$.

Baselines and S2N models are computed by following steps:
\begin{itemize}
    \item GLASS \& Connected: $\hmH = \mR^\top \gnn(\mX, \mA)$ where $\mR \in \sR^{N \times M}$ is a readout matrix.
    \item S2N+0: $\hmH = \gnn_{\text{S2N}}(\hmX, \hmA)$ where $\hmX_{[i]} = \sum_{v \in \sV_i^{\sub}} \omega_{vi} \cdot \mX_{[v, :]}$.
    \item S2N+A: $\hmH = \gnn_{\text{S2N}}(\hmX, \hmA)$ where $\hmX_{[i]} = \sum_{v \in \sV_i^{\sub}} \omega_{vi} \cdot \gnn_{\node} ( \mX_{[\sV_i^{\sub}, :]}, \mA_{i}^{\sub})_{[v, :]}$.
\end{itemize}
Graph neural networks (GNNs) that use the message-passing mechanism to learn subgraph representations can be decomposed into feature transformation (FT), feature propagation (FP), and subgraph-level readout (SR). Feature transformation requires $O(\text{the number of nodes} \times F^2)$ computations and feature propagation by sparse implementation requires $O(\text{the number of edges} \times F)$ computations. Plus, for the readout of representations or input features, we need the computations of $O(\text{the total number of nodes in subgraphs} \times F)$.
\begin{itemize}
    \item GLASS \& Connected: $O( EF )$ from FP, $O(M \overline{N^\sub} F)$ from SR, and $O(N F^2)$ from FT.
    \item GLASS: $O(M \overline{N^\sub})$ from the node labeling trick~\citep{wang2022glass}.
    \item S2N+0: $O( \hE F )$ from FP, $O(M \overline{N^\sub} F)$ from SR, and $O(M F^2)$ from FT.
    \item S2N+A: $O( \hE F )$ and $O( M \overline{E^\sub} F )$ from FP in $\gnn_{\text{S2N}}$ and $\gnn_{\text{node}}$, $O(M \overline{N^\sub} F)$ from SR, and $O(M F^2)$ and $O(M \overline{N^\sub} F^2)$ from FT in $\gnn_{\text{S2N}}$ and $\gnn_{\text{node}}$.
\end{itemize}
By adding up all the terms, we can get the final result.
\end{proof}

\begin{proposition}[Proposition~\ref{proposition:edge_cm}]
\propositionEdgeCMContent
\end{proposition}
\begin{proof}
We first note that the probability of edge $(u, v)$ in the Configuration Model for large $E$ is $\frac{d_u d_v}{2E}$ and $E = \frac{1}{2} \sum_k d_k = \frac{1}{2} N \E[d]$~\citep{newman2018networks}.
\begin{align}
P(\hmA_{[i,j]} \geq c)
&\leq \frac{\E [ \hmA_{[i,j]} ]}{c} \quad (\because \text{Markov's inequality}) \\
&= \frac{\E [ \sum_{u \in \sV_i^{\sub}} \sum_{v \in \sV_j^{\sub}} \mA_{[u, v]} ]}{c} \\
&= \frac{\E [ \sum_{u \in \sV_i^{\sub}} \sum_{v \in \sV_j^{\sub}} \frac{d_u d_v}{2E} ]}{c} \\
&= \frac{\E_{(i, j) \in \sS \times \sS} [ \sum_{u \in \sV_i^{\sub}} \sum_{v \in \sV_j^{\sub}} \E [d]^2 ]}{2cE} \\
&= \frac{ (\overline{N^\sub} \E[d] )^2}{2cE} \\
&= \frac{(\overline{N^\sub})^2 \E[d] }{cN}
\end{align}
\end{proof}

To prove Proposition~\ref{proposition:single_gcn_approx}, we first introduce Lemma~\ref{lemma:degree}.
\begin{lemma}\label{lemma:degree}
\begin{equation}
\mR^{\top} \mD^{-\frac{1}{2}} \mA \mD^{-\frac{1}{2}} \mR
= \hmD^{-\frac{1}{2}} \hmA \hmD^{-\frac{1}{2}}
\end{equation}
\end{lemma}
\begin{proof}
\begin{align}
\mR^{\top} \mD^{-\frac{1}{2}} \mA \mD^{-\frac{1}{2}} \mR 
&= (\mD^{\frac{1}{2}} \mM \hmD^{-\frac{1}{2}})^{\top} \mD^{-\frac{1}{2}} \mA \mD^{-\frac{1}{2}} \mD^{\frac{1}{2}} \mM \hmD^{-\frac{1}{2}} \\
&= \hmD^{-\frac{1}{2}} \mM^{\top} \mD^{\frac{1}{2}} \mD^{-\frac{1}{2}} \mA \mD^{-\frac{1}{2}} \mD^{\frac{1}{2}} \mM \hmD^{-\frac{1}{2}} \\
&= \hmD^{-\frac{1}{2}} \mM^{\top} \mA \mM \hmD^{-\frac{1}{2}} \\
&= \hmD^{-\frac{1}{2}} \hmA \hmD^{-\frac{1}{2}}.
\end{align}
\end{proof}

\begin{proposition}[Proposition~\ref{proposition:single_gcn_approx}]
\propositionSingleGCNApproxContent
\end{proposition}
\begin{proof}
\begin{align}
\lVert \mR^{\top}\mH - \hmH \rVert 
    &=
\lVert
    \mR^{\top} \mD^{-\frac{1}{2}} \mA \mD^{-\frac{1}{2}} \mX \mW
    - \hmD^{-\frac{1}{2}} \hmA \hmD^{-\frac{1}{2}} \hmX \mW
\rVert \\
    &=
\lVert
    \mR^{\top} \mD^{-\frac{1}{2}} \mA \mD^{-\frac{1}{2}} \mX \mW
    - \mR^{\top} \mD^{-\frac{1}{2}} \mA \mD^{-\frac{1}{2}} \mR \hmX \mW
\rVert\quad (\because \text{Lemma~\ref{lemma:degree}}) \\
    &=
\lVert
    \mR^{\top} (\mD^{-\frac{1}{2}} \mA \mD^{-\frac{1}{2}})
    ( \mX - \mR \hmX )  \mW
\rVert \\
    &\leq
\lVert
    \mR^{\top}
    \rVert
    \lVert
        \mD^{-\frac{1}{2}} \mA \mD^{-\frac{1}{2}}
    \rVert
    \lVert
        \mX - \mR \hmX
    \rVert 
    \lVert \mW
\rVert \\
    &\leq
M^{\frac{1}{2}} \lVert \mX - \mR \hmX \rVert
    \cdot
\lVert \mW \rVert.
\end{align}
\end{proof}

Although Proposition~\ref{proposition:single_gcn_approx} is analyzed using GCN models only, it is not limited to GCN in its applicability. Intuitively, when sufficient subgraph samples are unavailable, message-passing in any GNNs fails in the global graph not covered by existing subgraphs. Moreover, we can obtain theoretical results similar to Proposition~\ref{proposition:single_gcn_approx} for other GNNs. However, we might not get the approximation bound analytically depending on GNN architectures. For Graph Isomorphism Network (GIN)~\citep{xu2018how} as an example, the non-linearity in multi-layer perceptron (MLP) makes it hard to analytically compare the GIN outputs of S2N and the original graph. Instead, we introduce an approximation error bound on `GIN Sum-1-Layer', a less powerful variant of GINs that replaces MLP with single-layer perceptron (SLP).
\begin{align}
\text{GIN: }& \mH = \MLP \left( \left( \mathbf{A} +
(1 + \epsilon) \cdot \mathbf{I} \right) \cdot \mX \right), \\
\text{GIN Sum-1-Layer: }& \mH = \mathrm{SLP} \left( \left( \mathbf{A} +
(1 + \epsilon) \cdot \mathbf{I} \right) \cdot \mX \right).
\end{align}

The error bound between S2N's node representations and the global graph's subgraph representations is demonstrated in Proposition~\ref{proposition:single_gins1l_approx}. Here, we use the sum-readout $\readout(\mH) = \mM^{\top} \mH$ to get subgraph representations.

\begin{proposition}\label{proposition:single_gins1l_approx}
Using the single-layer GIN Sum-1-Layer parametrized by $\mW$, subgraph representations $\mM^{\top}\mH$ of the global graph $\gG$ can be approximated by node representations $\hmH$ of the S2N graph $\hgG$, that is, $\hmH \approx \mM^{\top}\mH$. The error between two representations is bounded by:
\begin{align}\label{eq:single_gins1l_approx}
\lVert \mM^{\top}\mH - \hmH \rVert
    \leq
    \left(
        (M \overline{N^{\sub}} E)^{\frac{1}{2}}
        \lVert \mX - \mM \hmX \rVert
        + (1+\eps)
        \lVert \mM^{\top} \mX -  \hmX \rVert 
    \right)
    \cdot \lVert \mW
\rVert.
\end{align}
\end{proposition}
\begin{proof}
\begin{align}
\lVert \mM^{\top}\mH - \hmH \rVert 
    &=
\lVert
    \mM^{\top} (\mA +  (1+\eps) \mI_{N}) \mX \mW
    - (\hmA +  (1+\eps) \mI_{M}) \hmX \mW
\rVert \\
    &=
\lVert
    \mM^{\top} (\mA +  (1+\eps) \mI_{N}) \mX \mW
    - (\mM^{\top} \mA \mM+  (1+\eps) \mI_{M}) \hmX \mW
\rVert \\
    &=
\lVert
    \mM^{\top} \mA (\mX - \mM \hmX) \mX \mW
    +  (1+\eps) (\mM^{\top} \mX - \hmX)  \mW
\rVert \\
    &\leq
\lVert
    \mM^{\top}
    \rVert
    \lVert \mA \rVert
    \lVert \mX - \mM \hmX \rVert 
    \lVert \mW \rVert
    + (1+\eps)
    \lVert \mM^{\top} \mX -  \hmX \rVert 
    \lVert \mW
\rVert \\
    &\leq
    \left(
        (M \overline{N^{\sub}} E)^{\frac{1}{2}}
        \lVert \mX - \mM \hmX \rVert
        + (1+\eps)
        \lVert \mM^{\top} \mX -  \hmX \rVert 
    \right)
    \cdot \lVert \mW
\rVert,
\end{align}
where $\mI_{N}$ is an identity matrix of size $N$.
\end{proof}
If we set the initial features of S2N as a sum of the original features (i.e., $\hmX = \mM^{\top} \mX$), Corollary~\ref{corollary:simple_single_gins1l_approx} then follows from Proposition~\ref{proposition:single_gins1l_approx}.
\begin{corollary}\label{corollary:simple_single_gins1l_approx}
Using the single-layer GIN Sum-1-Layer parametrized by $\mW$, subgraph representations $\mM^{\top}\mH$ of the global graph $\gG$ can be approximated by node representations $\hmH$ of the S2N graph $\hgG$, that is, $\hmH \approx \mM^{\top}\mH$. If the initial feature matrix of S2N is $\hmX = \mM^{\top} \mX$, the error between two representations is bounded by:
\begin{align}\label{eq:simple_single_gins1l_approx}
\lVert \mM^{\top}\mH - \hmH \rVert
    \leq
    (M \overline{N^{\sub}} E)^{\frac{1}{2}}
    \lVert
        \mX - \mM \hmX
    \rVert
        \cdot
    \lVert \mW \rVert.
\end{align}
\end{corollary}

\begin{table}[t]
\centering
\caption{Statistics of real-world datasets in original forms (before S2N translation).}
\label{tab:dataset_original}
\begin{tabular}{lllll}
\hline
                               & \PPIBP   & \HPONeuro & \HPOMetab & \EMUser \\ \hline
\# nodes in $\gG$  & 17,080    & 14,587    & 14,587    & 57,333    \\
\# edges in $\gG$  & 316,951   & 3,238,174 & 3,238,174 & 4,573,417 \\
\# internal edges in subgraphs & 9,627 & 217,555 & 390,450 & 86,648 \\
\# subgraphs                   & 1,591    & 4,000     & 2,400     & 324       \\ \hline
Density of $\gG$ & 0.0022      & 0.0304    & 0.0304    & 0.0028    \\ 
Average density of subgraphs      & 0.216$_{\pm 0.188}$ & 0.767$_{\pm 0.141}$ & 0.757$_{\pm 0.149}$ & 0.010$_{\pm 0.006}$ \\ \hline
Average \# nodes / subgraph      & 10.2$_{\pm 10.5}$   & 14.8$_{\pm 6.5}$    & 14.4$_{\pm 6.2}$    & 155.4$_{\pm 100.2}$ \\
Average \# components / subgraph & 7.0$_{\pm 5.5}$     & 1.5$_{\pm 0.7}$     & 1.6$_{\pm 0.7}$     & 52.1$_{\pm 15.3}$   \\ \hline
\# classes                  & 6        & 10        & 6         & 2         \\
Single- or multi-label           & Single-label       & Multi-label        & Single-label       & Single-label       \\ \hline
Train/Valid/Test splits     & 80/10/10 & 80/10/10  & 80/10/10  & 70/15/15  \\ \hline
\end{tabular}
\end{table}

\section{Datasets}\label{appendix:dataset}

All real-world subgraph datasets (\PPIBP, \HPONeuro, \HPOMetab, and \EMUser) and synthetic subgraph datasets (\Density, \CutRatio, \Coreness, and \Component) are proposed in \citet{alsentzer2020subgraph}. They can be downloaded from the author's GitHub repository\footnote{\url{https://github.com/mims-harvard/SubGNN}}. Pre-trained embeddings can be downloaded from the GitHub repository\footnote{\url{https://github.com/Xi-yuanWang/GLASS}} of \citet{wang2022glass}. The following paragraphs describe their nodes, edges, subgraphs, tasks, and references. Note that the number of edges in the real-world datasets compared to datasets referred to as large-scale~\citep{lim2021large} is at a similar level; thus, similar scalability is required to model real-world graphs using GNNs.

\subsection{Real-World Datasets}

\paragraph{\PPIBP}

The global graph of \PPIBP~\citep{zitnik2018biosnap, subramanian2005gene, gene2019gene, ashburner2000gene} is a human protein-protein interaction (PPI) network; nodes are proteins, and edges are whether there is a physical interaction between proteins. Subgraphs are sets of proteins in the same biological process (e.g., alcohol bio-synthetic process). The task is to classify processes into six categories.

\paragraph{\HPONeurob and \HPOMetab}

These two HPO (Human Phenotype Ontology) datasets~\citep{hartley2020new, kohler2019expansion, mordaunt2020metabolomics} are knowledge graphs of phenotypes (i.e., symptoms) of rare neurological and metabolic diseases. Each subgraph is a collection of symptoms associated with a monogenic disorder. The task is to diagnose the rare disease: classifying the disease type among subcategories (ten for \HPONeurob and six for \HPOMetab).

\paragraph{\EMUser}

\EMUserb (Users in EndoMondo) dataset is a social fitness network from Endomondo~\citep{ni2019endomondo}. Here, subgraphs are users, nodes are workouts, and edges exist between workouts completed by multiple users. Each subgraph represents the workout history of a user. The task is to profile a user's gender.

\subsection{Synthetic Datasets}

\paragraph{\Density, \CutRatio, \Coreness, and \Component}

\begin{table}[t]
\centering
\caption{Statistics of synthetic datasets in original forms (before S2N translation).}
\label{tab:dataset_original_syn}
\begin{tabular}{lllll}
\hline
                   & \Density   & \CutRatio & \Coreness & \Component \\ \hline
\# nodes in $\gG$  & 5,000    & 5,000    & 5,000    & 19,555     \\
\# edges in $\gG$  & 29,521    & 83,969 & 118,785 & 43,701 \\
\# subgraphs       & 250    & 250    & 221     & 250       \\ \hline
Density of $\gG$   & 0.0024       & 0.0067    & 0.0095    & 0.0002    \\ 
Average density of subgraphs      & 0.232$_{\pm 0.146}$ & 0.945$_{\pm 0.028}$ & 0.219$_{\pm 0.062}$ & 0.150$_{\pm 0.161}$ \\ \hline
Average \# nodes / subgraph      & 20.0$_{\pm 0.0}$   & 20.0$_{\pm 0.0}$    & 20.0$_{\pm 0.0}$    & 74.2$_{\pm 52.8}$ \\
Average \# components / subgraph & 3.8$_{\pm 3.7}$     & 1.0$_{\pm 0.0}$     & 1.0$_{\pm 0.0}$     & 4.9$_{\pm 3.5}$   \\ \hline
\# classes                  & 3        & 3        & 3         & 2         \\
Single- or multi-label           & Single-label       & Single-label        & Single-label       & Single-label       \\ \hline
Train/Valid/Test splits     & 80/10/10 & 80/10/10  & 80/10/10  & 80/10/10  \\ \hline
\end{tabular}
\end{table}

\begin{table}[t]
\centering
\caption{The attributes that affect the subgraph properties (labels) of synthetic datasets, introduced in \citet{alsentzer2020subgraph}.}
\label{tab:syn_properties}
\resizebox{\textwidth}{!}{%
\begin{tabular}{llll}
\hline
\Density           & \CutRatio        & \Coreness                                         & \Component                     \\ \hline
Internal structure & Border structure & Internal structure, border structure \& position & Internal \& external position \\ \hline
\end{tabular}%
}
\end{table}

For these synthetic datasets, the task is to predict the properties of subgraphs: density, cut ratio, average core number, and the number of components, respectively. Refer to \citet{alsentzer2020subgraph} for details on how to generate the synthetic graphs. We use a vector of 64 dimensions initialized to 1 or its L1-normalized vector as input node embedding.

\section{Models}\label{appendix:model}

We describe the hyperparameter details and the tuning method. All models are implemented with PyTorch~\citep{paszke2019pytorch}, PyTorch Geometric~\citep{Fey2019Fast}, and PyTorch Lightning~\citep{Falcon_PyTorch_Lightning_2019}.

We tune hyperparameters using TPE (Tree-structured Parzen Estimator) algorithm in Optuna~\citep{akiba2019optuna} by 400 trials: learning rate ($5 \times 10^{-4}$ -- $10^{-2}$), weight decay ($10^{-9}$ -- $10^{-6}$), the number of layers in $\gnn$ (1 -- 2), dropout of channels and edges ($\{ 0.0, 0.1, ..., 0.5 \}$), gradient clipping ($\{ 0.0, 0.1, ..., 0.5 \}$), the readout matrix ($\omega_{vi} = \mR_{[v, i]}$ in Equation~\ref{eq:readout_matrix} or $\omega_{vi} = \mM_{[v, i]}$), and whether to use batch normalization~\citep{ioffe2015batch} and skip-connection~\citep{he2016deep}. Hyperparameters specialized on GCNII are also tuned: $\alpha$ ($\{ 0.1, 0.2, ..., 0.9 \}$), $\theta$ ($\{ 0.1, 0.2, ..., 2.0 \}$), weight sharing (True or False). For S2N translation, we tune edge normalization range ($a$ and $b = a + \Delta$ in Equation~\ref{eq:s2n_edge_normalization}, $a \in \{ 1.0, 1.25, ..., 4.0 \}$, $\Delta \in \{ 0.5, 1.0, 1.5, 2.0 \}$). We add frozen Random Walk Positional Encoding (RWPE)~\citep{dwivedi2022graph} to input features for real-world datasets. For synthetic datasets, we allocate RWPE to 1/2 or 1/4 of the total embedding dimension.

All hyperparameters are reported in the code.

\section{Efficiency Measurement}\label{appendix:efficiency}

We compute throughput (subgraphs per second) and latency (seconds per forward pass) using the following equations. In addition, we use \texttt{torch.cuda.max\_memory\_allocated} to measure the maximum allocated GPU VRAM\footnote{\href{https://pytorch.org/docs/1.9.0/generated/torch.cuda.max\_memory\_allocated.html}{https://pytorch.org/docs/1.9.0/generated/torch.cuda.max\_memory\_allocated.html}}.
\begin{align}
\text{Training throughput} &= \frac{\text{\# of training subgraphs}}{\text{training wall-clock time (seconds) / \# of epochs}}, \\
\text{Evaluation throughput} &= \frac{\text{\# of validation subgraphs}}{\text{validation wall-clock time (seconds) / \# of epochs}}, \\
\text{Training latency} &= \frac{\text{training wall-clock time (seconds)}}{\text{\# of training batches}}, \\
\text{Evaluation latency} &= \frac{\text{validation wall-clock time (seconds)}}{\text{\# of validation batches}}.
\end{align}

While throughput is a primary metric in practice, focusing solely on this metric is suboptimal for best model selection. Specifically, we observed that connected forms suffer from extensive memory consumption, and separated forms exhibit significant performance degradation compared to our proposed S2N models. The holistic evaluation of performance, throughput, and memory requirements is crucial for practitioners to decide which models to employ based on their specific constraints and requirements, as stated in \citet{dehghani2022the}.

\section{Discussion on the Number of Nodes and Edges in S2N}\label{appendix:number_of_nodes}

The number of nodes in S2N+A in Table~\ref{tab:s2n_statistics} only includes nodes translated from subgraphs. The total number of nodes in S2N+A may be larger than the original global graph if we count all internal nodes in the subgraph. However, this is not significantly related to actual efficiency. The reason is that in S2N+A, internal nodes in the subgraph are kept sparse (i.e., indexing) rather than dense embedding by implementation, so memory cost is low. Specifically, node indexing requires the space complexity of $O(N)$, and dense embedding requires $O(NF)$, where $N$ is the number of nodes and $F$ is the number of features. The dominant factors of computational and memory bottlenecks are the number of subgraphs (the number of nodes in S2N), which determines the size of the final representation, and the number of edges, which determines the number of message-passing, as illustrated in Table~\ref{tab:s2n_statistics}.

\section{Generalization of Homophily to Multi-label Classification}\label{appendix:homophily_ml}

Node~\citep{Pei2020Geom-GCN:} and edge homophily~\citep{zhu2020beyond} are defined by,
\begin{gather}
h^{\text{edge}}
= \frac{| \{ (u, v) | (u, v) \in \sA \wedge y_u = y_v \} | }{ |\sA| },\
h^{\text{node}}
= \frac{1}{|\sV|} \sum_{v \in \sV} \frac{ | \{ (u, v) | u \in \mathcal{N}(v) \wedge y_u = y_v \} |  } { |\mathcal{N}(v)| },
\end{gather}
where $y_v$ is the label of the node $v$. In the main paper, we define multi-label node and edge homophily by,
\begin{gather}
h^{\text{edge, ml}}
= \frac{ 1 }{ |\sA| } \sum_{ (u, v) \in \sA } \frac{| \sL_u \cap \sL_v |}{| \sL_u \cup \sL_v |},\
h^{\text{node, ml}}
= \frac{1}{|\sV|} \sum_{v \in \sV} \left(
    \frac{ 1 } { |\mathcal{N}(v)| } \sum_{ u \in \mathcal{N}(v) } \frac{| \sL_u \cap \sL_v |}{| \sL_u \cup \sL_v |}
\right).
\end{gather}
If we compute $r = \frac{| \sL_u \cap \sL_v |}{| \sL_u \cup \sL_v |}$ for single-label multi-class graphs, $r = \frac{1}{1} = 1$ for nodes of same classes, and $r = \frac{0}{2} = 0$ for nodes of different classes. That makes $h^{\text{edge, ml}} = h^{\text{edge}}$ and $h^{\text{node, ml}} = h^{\text{node}}$ for single-label graphs.

\begin{table}[]
\centering
\caption{Mean performance in micro F1-score over 10 runs. We mark with daggers the reprinted results from \citet{alsentzer2020subgraph} ($\dagger$) and \citet{wang2022glass} ($\ddagger$).}
\label{tab:results_diff_gnn}
\begin{tabular}{llll}
\hline
Model                  & Data Structure & \PPIBP           & \EMUser          \\ \hline
Sub2Vec Best$^\dagger$ &                & $30.9_{\pm 2.3}$ & $85.9_{\pm 1.4}$ \\
SubGNN$^\dagger$       &                & $59.9_{\pm 2.4}$ & $81.4_{\pm 4.6}$ \\
GLASS$^\ddagger$       &                & $61.9_{\pm 0.7}$ & $88.8_{\pm 0.6}$ \\ \hline
GCNII                  & Separated      & $61.3_{\pm 1.2}$ & $84.7_{\pm 4.1}$ \\
GCNII                  & Connected      & $63.5_{\pm 2.0}$ & $85.5_{\pm 4.8}$ \\
GCNII                  & S2N+0          & $63.5_{\pm 2.4}$ & $86.5_{\pm 3.2}$ \\
GCNII                  & S2N+A          & $63.7_{\pm 2.3}$ & $89.0_{\pm 1.6}$ \\ \hline
GIN                    & Separated      & $60.6_{\pm 2.1}$ & $82.2_{\pm 6.6}$ \\
GIN                    & Connected      & $61.0_{\pm 3.3}$ & $83.7_{\pm 4.8}$ \\
GIN                    & S2N+0          & $63.3_{\pm 1.6}$ & $84.9_{\pm 5.3}$ \\
GIN                    & S2N+A          & $62.2_{\pm 1.9}$ & $83.1_{\pm 1.6}$ \\ \hline
GATv2                  & Separated      & $61.4_{\pm 2.6}$ & $84.7_{\pm 4.9}$ \\
GATv2                  & Connected      & $61.0_{\pm 1.5}$ & OOM              \\
GATv2                  & S2N+0          & $62.8_{\pm 1.7}$ & $84.9_{\pm 2.4}$ \\
GATv2                  & S2N+A          & $62.6_{\pm 1.4}$ & $86.7_{\pm 3.2}$ \\ \hline
\end{tabular}
\end{table}
\begin{table}[t]
\centering
\caption{Mean performance in micro F1-score over 10 runs using GCNII models with different readout methods.}
\label{tab:ablation_readout}
\begin{tabular}{llllll}
\hline
Data Structure         & Readout & \PPIBP                    & \HPONeuro                 & \HPOMetab                 & \EMUser                   \\ \hline
\multirow{4}{*}{S2N+0} & Sum     & $\mathbf{63.5_{\pm 2.4}}$ & $\mathbf{66.4_{\pm 1.1}}$  & $61.6_{\pm 1.7}$ & $\mathbf{86.5_{\pm 3.2}}$ \\
                       & Mean    & $59.9_{\pm 2.0}$          & $63.9_{\pm 0.5}$          & $\mathbf{62.0_{\pm 1.0}}$          & $85.3_{\pm 3.9}$          \\
                       & Max     & $48.6_{\pm 2.8}$          & $54.1_{\pm 1.0}$          & $51.8_{\pm 2.2}$          & $72.2_{\pm 7.1}$          \\
                       & Degree  & $60.3_{\pm 2.2}$          & $65.6_{\pm 1.0}$          & $58.8_{\pm 2.3}$          & $83.1_{\pm 2.9}$           \\ \hline
\multirow{4}{*}{S2N+A} & Sum     & $\mathbf{63.7_{\pm 2.3}}$ & $\mathbf{68.4_{\pm 1.0}}$ & $\mathbf{63.2_{\pm 2.7}}$ & $88.8_{\pm 2.1}$          \\
                       & Mean    & $59.6_{\pm 1.5}$          & $66.6_{\pm 1.0}$          & $60.8_{\pm 1.1}$          & $88.0_{\pm 3.2}$          \\
                       & Max     & $58.5_{\pm 1.7}$          & $59.6_{\pm 2.0}$          & $59.4_{\pm 2.7}$          & $81.2_{\pm 3.1}$          \\
                       & Degree  & $63.1_{\pm 2.2}$          & $\mathbf{68.4_{\pm 0.9}}$          & $61.0_{\pm 2.0}$          & $\mathbf{89.0_{\pm 1.6}}$ \\ \hline
\end{tabular}
\end{table}

\begin{table}[t]
\centering
\caption{Mean performance in micro F1-score over 10 runs using GCNII models with different numbers of layers of $\gnn_{\STON}$.}
\label{tab:ablation_num_layers}
\begin{tabular}{llllll}
\hline
Data Structure         & \# layers & \PPIBP                    & \HPONeuro                 & \HPOMetab                 & \EMUser                   \\ \hline
\multirow{4}{*}{S2N+0} & 0         & $57.7_{\pm 1.6}$          & $65.2_{\pm 1.6}$          & $55.5_{\pm 1.9}$          & $77.6_{\pm 9.4}$          \\
                       & 1         & $61.1_{\pm 2.4}$          & $\mathbf{66.4_{\pm 1.1}}$ & $\mathbf{61.6_{\pm 1.7}}$ & $79.2_{\pm 9.2}$          \\
                       & 2         & $\mathbf{63.5_{\pm 2.4}}$ & $65.6_{\pm 1.4}$          & $59.4_{\pm 1.0}$          & $\mathbf{86.5_{\pm 3.2}}$ \\
                       & 4         & $62.8_{\pm 2.0}$          & $65.8_{\pm 0.8}$          & $61.1_{\pm 1.7}$          & $79.2_{\pm 7.9}$          \\ \hline
\multirow{4}{*}{S2N+A} & 0         & $59.7_{\pm 2.2}$          & $68.2_{\pm 0.8}$          & $61.8_{\pm 1.7}$          & $87.1_{\pm 3.5}$          \\
                       & 1         & $\mathbf{63.7_{\pm 2.3}}$ & $\mathbf{68.4_{\pm 1.0}}$ & $61.9_{\pm 2.0}$          & $\mathbf{89.0_{\pm 1.6}}$ \\
                       & 2         & $61.8_{\pm 1.4}$          & $68.0_{\pm 0.8}$          & $\mathbf{63.2_{\pm 2.7}}$ & $86.3_{\pm 4.9}$          \\
                       & 4         & $61.6_{\pm 1.7}$          & $67.7_{\pm 0.8}$          & $62.0_{\pm 1.6}$          & $86.3_{\pm 5.2}$          \\ \hline
\end{tabular}
\end{table}

\begin{table}[t]
\centering
\caption{Mean performance in micro F1-score over 10 runs using GCNII models with different positional encoding.}
\label{tab:ablation_pe}
\begin{tabular}{llllll}
\hline
Data Structure         & Positional Encoding & \PPIBP                    & \HPONeuro                 & \HPOMetab                 & \EMUser                   \\ \hline
\multirow{3}{*}{S2N+0} & None                & $63.5_{\pm 2.4}$          & $66.4_{\pm 1.1}$          & $61.6_{\pm 1.7}$          & $86.5_{\pm 3.2}$          \\
                       & RWPE                & $63.5_{\pm 1.7}$          & $\mathbf{66.7_{\pm 0.6}}$ & $\mathbf{62.3_{\pm 1.1}}$ & $86.5_{\pm 4.7}$          \\
                       & LapPE               & $\mathbf{63.9_{\pm 2.4}}$ & $66.1_{\pm 0.8}$          & $61.9_{\pm 1.7}$          & $\mathbf{87.3_{\pm 2.5}}$ \\ \hline
\multirow{3}{*}{S2N+A} & None                & $63.7_{\pm 2.3}$          & $68.4_{\pm 1.0}$          & $63.2_{\pm 2.7}$          & $\mathbf{89.0_{\pm 1.6}}$ \\
                       & RWPE                & $\mathbf{64.3_{\pm 1.8}}$ & $68.6_{\pm 0.8}$          & $\mathbf{63.9_{\pm 1.7}}$ & $\mathbf{89.0_{\pm 3.1}}$ \\
                       & LapPE               & $64.2_{\pm 1.9}$          & $\mathbf{68.8_{\pm 0.9}}$ & $63.7_{\pm 1.4}$          & $\mathbf{89.0_{\pm 3.2}}$ \\ \hline
\end{tabular}
\end{table}

\section{Performance of Different GNN Layers}\label{appendix:perf_of_diff_gnns}

In Table~\ref{tab:results_diff_gnn}, we demonstrate the performance of S2N models using additional GNN layers: Graph Isomorphism Networks (GIN)~\citep{xu2018how} and Graph Attention Networks v2 (GATv2)~\citep{brody2022attentive} on \PPIBPb and \EMUser.

GIN and GATv2 (S2N+0 and S2N+A) outperform GLASS on \PPIBPb but perform worse than on \EMUser. We confirm that S2N outperforms classic data structures: separated and connected forms. For GATv2, we cannot experiment with the connected form on \EMUserb due to the requirements of large GPU memory. Nonetheless, all S2N models with GIN and GATv2 outperform SubGNN on all datasets.

Compared to GCNII, which showed the best performance in our paper, GIN and GATv2 generally perform worse. This implies that architectures designed for node or link-level tasks are sub-optimal for subgraph-level tasks. We suggest further studies on model architectures for learning subgraph representations.

\section{Ablation Study of Hyperparameters}\label{appendix:ablation}

We conduct ablation studies on the readout method (Equation~\ref{eq:readout_function}) (sum, mean, max, and degree-dependent), the number of layers in $\gnn_{\STON}$ (0, 1, 2, 4), and the positional encoding~\citep{dwivedi2022graph}. We report the performance of S2N+0 and S2N+A with GCNII by the readout method in Table~\ref{tab:ablation_readout}, the number of layers in Table~\ref{tab:ablation_num_layers}, and the positional encoding in Table~\ref{tab:ablation_pe}.

\paragraph{Readout}

Generally, the sum-readout performs best, and the max-readout performs the worst, as illustrated in Table~\ref{tab:ablation_readout}. The performance of mean-readout and degree-dependent readout varies by dataset. In S2N+A, degree-dependent readout performs similarly to sum-readout and slightly outperforms on \EMUser.

\paragraph{The Number of Layers}

In Table~\ref{tab:ablation_num_layers}, we find that using message-passing (i.e., the number of layers $> 0$) always increases the performance on all datasets. That is, modeling the S2N graph structures helps to learn the representation of subgraphs. The performance improvement by $\gnn_{\STON}$ in S2N+0 is higher than in S2N+A, which leverages internal structures. The performance decreases when we use a deeper $\gnn_{\STON}$ than the optimum; that is, an over-smoothing effect exists in $\gnn_{\STON}$~\citep{li2018deeper}.

\paragraph{Positional Encoding}

In Table~\ref{tab:ablation_pe}, we report the performance of GCNII models with Random Walk Positional Encoding (RWPE) and Laplacian Positional Encoding (LapPE)~\citep{dwivedi2022graph}. We find that LapPE also contributes to performance improvement in general, and there is no significant difference between RWPE and LapPE.

\begin{figure*}[t]
  \centering
  \hspace{-1.4cm}
  \begin{subfigure}[b]{0.24\textwidth}
    \centering
    \includegraphics[height=0.145\textheight]{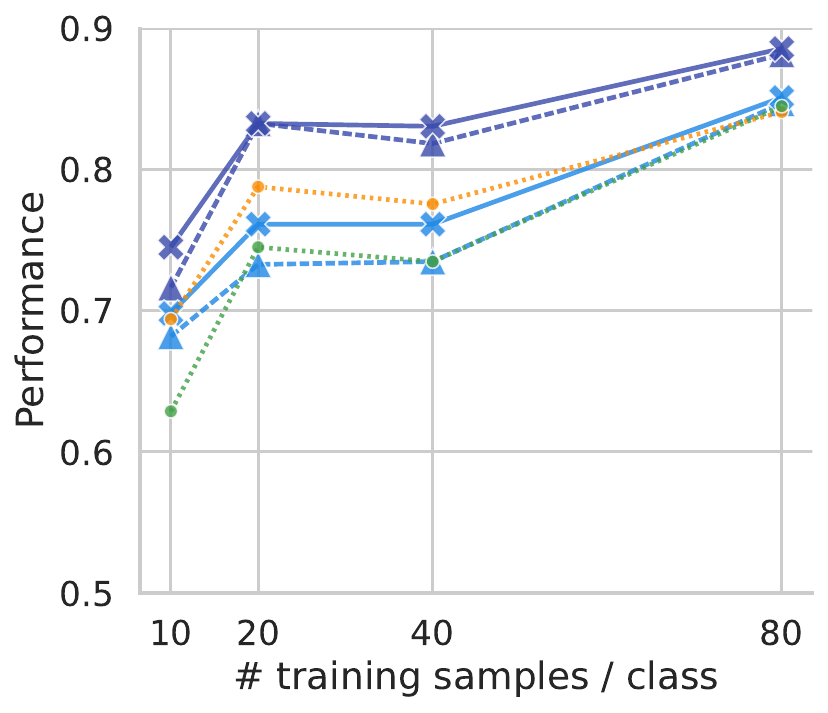}
    \caption{Performance}
  \end{subfigure}
  \hspace{-0.35cm}
  \begin{subfigure}[b]{0.24\textwidth}
    \centering
    \includegraphics[height=0.145\textheight]{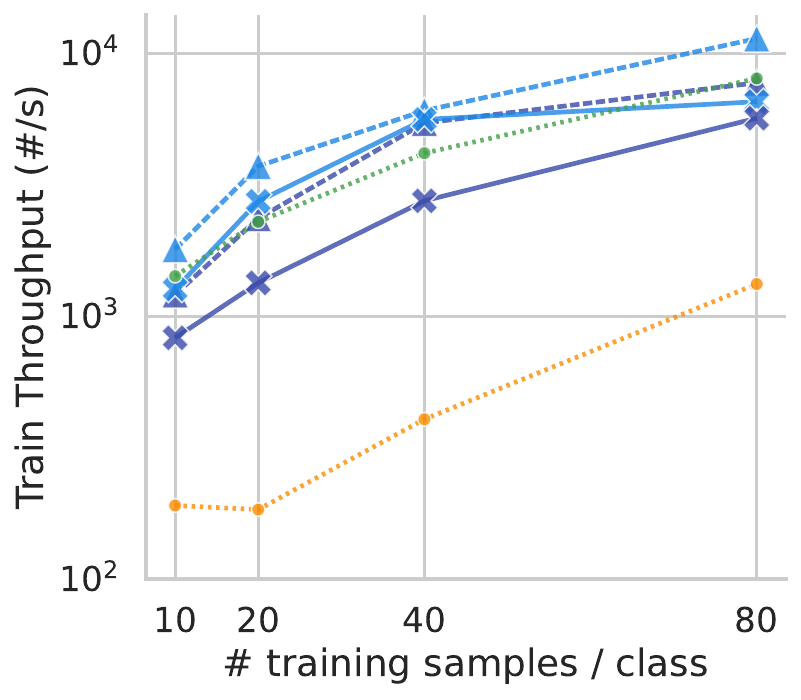}
    \caption{Training Throughput}
  \end{subfigure}
  \hspace{-0.35cm}
  \begin{subfigure}[b]{0.24\textwidth}
    \centering
    \includegraphics[height=0.145\textheight]{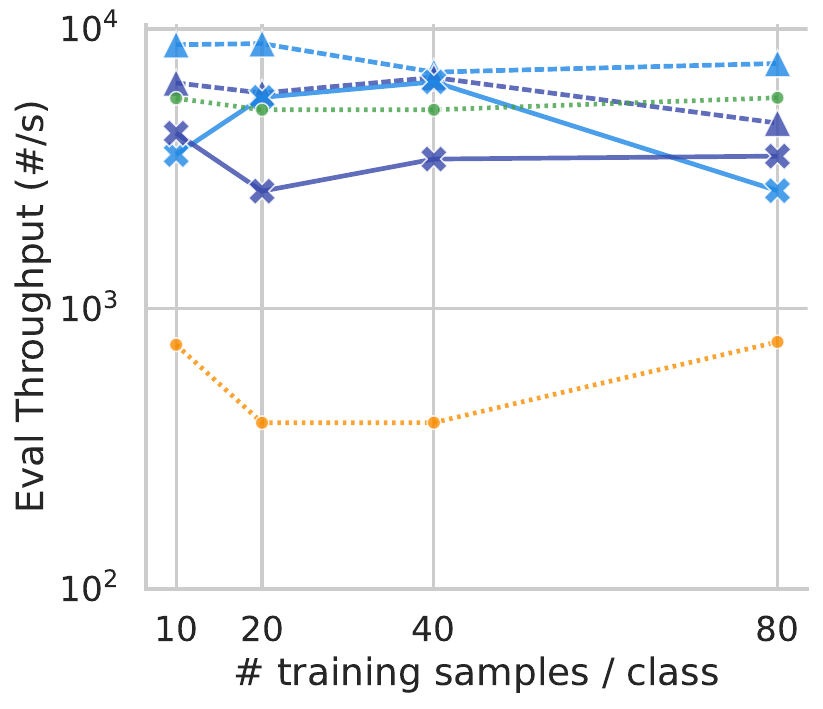}
    \caption{Eval Throughput}
  \end{subfigure}
  \begin{subfigure}[b]{0.24\textwidth}
    \centering
    \includegraphics[height=0.145\textheight]{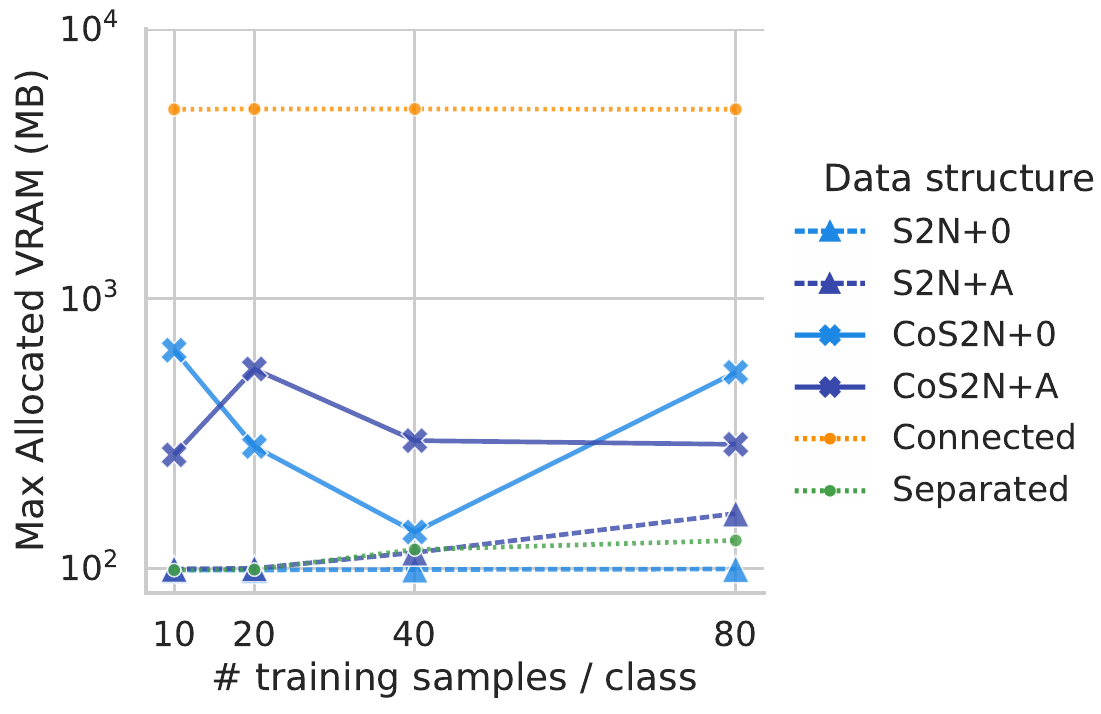}
    \caption{Max allocated VRAM}
  \end{subfigure}
  \caption{Performance and efficiency on \EMUserb of S2N, CoS2N, connected, and separated forms by the number of training samples in a data-scarce setting.}
  \label{fig:efficiency_by_num_training_em_user_3}
\end{figure*}

\begin{figure*}[t]
  \centering
  \begin{subfigure}[t]{\textwidth}
    \centering
    \includegraphics[width=\textwidth]{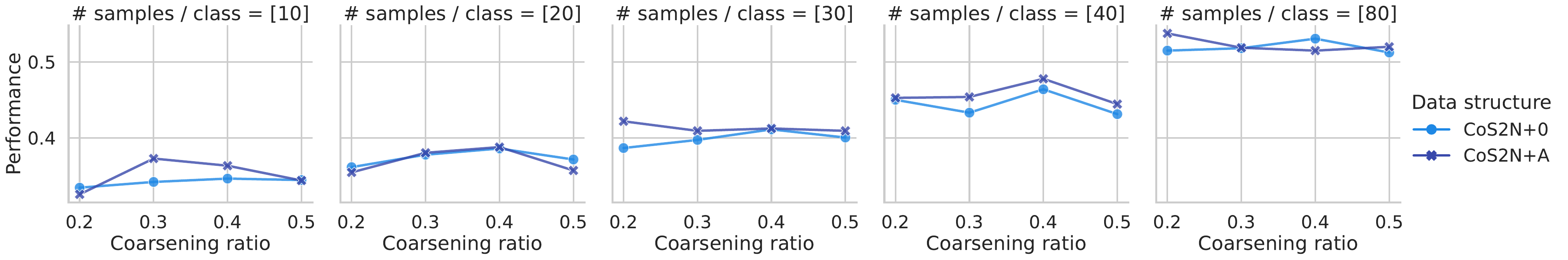}
    \vspace{-0.5cm}
    \caption{Performance on \PPIBPb by coarsening ratio.}
    \vspace{0.25cm}
  \end{subfigure}
  \begin{subfigure}[t]{\textwidth}
    \centering
    \includegraphics[width=\textwidth]{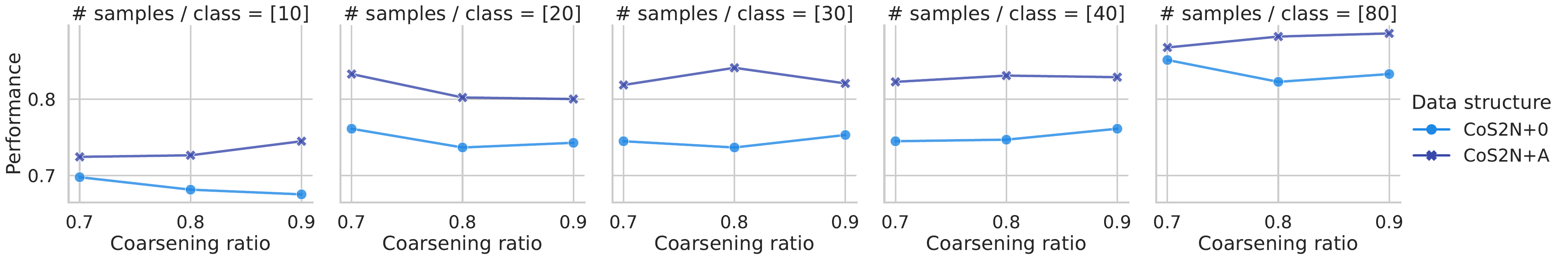}
    \vspace{-0.5cm}
    \caption{Performance on \EMUserb by coarsening ratio.}
    \vspace{0.25cm}
  \end{subfigure}
  \vspace{-0.3cm}
  \caption{Performance of CoS2N on \PPIBPb and \EMUserb by coarsening ratio.}
  \label{fig:performance_by_coarsening_ratio}
\end{figure*}

\section{Performance and Efficiency of Coarsened S2N in a Data-Scarce Setting}\label{appendix:efficiency_all_by_num_training}

For experiments in a data-scare setting, we narrow the search space of hyperparameters. Specifically, we fix to use batch normalization but not skip-connections. We use a coarsening ratio that creates virtual subgraphs smaller than the average size: $[0.2, 0.3, 0.4, 0.5]$ for \PPIBPb and $[0.7, 0.8, 0.9]$ for \EMUser. After graph coarsening, we remove subgraphs that consist of a single node. We follow the same tuning procedures in Appendix~\ref{appendix:model} for the remaining details. 

As stated in \S\ref{sec:result_efficiency_by_num_training}, we summarize performance and efficiency on \EMUserb in Figure~\ref{fig:efficiency_by_num_training_em_user_3}. Overall, results on \EMUserb do not show a notable difference from trends in data-scarce experiments on \PPIBPb at \S\ref{sec:result_efficiency_by_num_training}. We can observe that (1) subgraphs created by coarsening contribute to performance improvements of S2N, and (2) CoS2N has higher throughput and uses less memory than using the global graph.

We report the performance on \PPIBPb and \EMUserb by the coarsening ratio in Figure~\ref{fig:performance_by_coarsening_ratio}. Although there is no consistently optimal coarsening ratio by the number of training samples, we can conclude that finding the optimal coarsening ratio for each dataset can increase the performance.


\end{document}